%% file: iclr2020_conference.tex
\documentclass{article} % For LaTeX2e

\usepackage{iclr2019_conference,times}
\usepackage{graphicx,enumitem}
\usepackage[normalem]{ulem}
\usepackage[font=footnotesize,labelfont=bf]{caption}
\usepackage{graphicx}
\usepackage{caption}
\usepackage{subcaption}
\usepackage{fontawesome}

\useunder{\uline}{\ul}{}
\usepackage{nccmath}
\input{math_commands.tex}

\def \x {\mathbf{x}}

\def \x {\mathbf{x}}

\def \p {\mathbf{p}}

\def \P {\mathcal{P}}
\def \N {\mathcal{N}}

\def \q \textbf{q}
\usepackage{hyperref}
\usepackage{url}

\newtheorem{lem}{Lemma}

\newtheorem{remark}{Remark}

\title{A Simple and Effective Framework for Pairwise Deep Metric Learning}

\author{%
Qi Qi$^\dagger$, Yan Yan$^\dagger$, Xiaoyu Wang$^\ddagger$, Tianbao Yang\textsuperscript{\faEnvelopeO} $^\dagger$\\
$^\dagger$ Department of Computer Science, The University of Iowa, Iowa City, IA 52242, USA \\
  $^\ddagger$ The Chinese University of Hong Kong (Shenzhen), Shenzhen, China\\
  \texttt{\{qi-qi,yan-yan-2,tianbao-yang\}@uiowa.edu,fanghuaxue@gmail.edu}\\
}

\iclrfinalcopy % Uncomment for camera-ready version, but NOT for submission.
\begin{document}
\maketitle

$$\text{First version: December 24, 2019}$$ 
\vspace{-0.2in}
$$\text{Revised version: June 18, 2020}$$

\begin{abstract}
Deep metric learning (DML) has received much attention in deep learning due to its wide applications in computer vision.  
Previous studies have focused on designing complicated losses and hard example mining methods, which are mostly heuristic and lack of theoretical understanding. In this paper, we cast DML as a simple pairwise binary classification problem that classifies a pair of examples as similar or dissimilar. It identifies the most critical issue in this problem---imbalanced data pairs. 
To tackle this issue, we propose a simple and effective framework to sample pairs in a batch of data for updating the model.
The key to this framework is to define a robust loss for all pairs over a mini-batch of data, which is formulated by distributionally robust optimization.
The flexibility in constructing the  {\it uncertainty decision set} of the dual variable allows us to recover state-of-the-art complicated losses and also to induce novel variants. 
Empirical studies on several benchmark data sets demonstrate that our simple and effective method outperforms the state-of-the-art results \footnote{Codes are available online at \url{https://github.com/qiqi-helloworld/A-Simple-and-Effective-Framework-for-Pairewise-Distance-Metric-Learning}}

\setlength{\abovedisplayskip}{3pt}
\setlength{\belowdisplayskip}{3pt}

%Pairwise losses have been shown outstanding performance on learning a good embedding in Distance Metric Learning (DML). Pairs are constructed within a batch in the training process. The most widely used empirical batch construction methods in DML randomly sample a certain number of classes with a certain number samples for each class. By simple calculation, it is easy to show that positive pairs and negative pairs are highly imbalanced in a batch. To address this issue, previous sampling methods simply down-sampling the negative pairs with some strategies so that the number of negative samples is the same as the positive samples, which ignores the intrinsic property of the imbalance problem: minor classes do not have equal contribution to the loss objective. In this work, we propose a unified framework that minimizing the top-K losses in a batch. Further more, our framework are able to cover different state-of-art losses. Experiment shows ...
\end{abstract}

\section{Introduction}
\vspace*{-0.1in}\label{sec:introduction}
Metric Learning aims to learn a metric to measure the distance between examples that captures certain notion of human-defined similarity between examples. Deep metric learning (DML) has emerged as an effective approach for learning a metric by training a deep neural network. Simply speaking, a deep neural network can induce new feature embedding of examples and it is trained in such a way that the Euclidean distance between the induced feature embeddings of two similar examples shall be small and that between the induced feature embeddings of two dissimilar pairs shall be large. DML has been widely used in many tasks such as face recognition~(\cite{fan2017learning}), image retrieval~(\cite{chen2019hybrid}), and classification~(\cite{Qian_2015_CVPR,li2019revisiting}). 

%can be seen as a pre-requisite to achieve high performance in many machine learning tasks.  Recently, deep learning has shown its dominant representation ability on images and videos. Equipped with deep neural networks, deep metric learning (DML) receives great prevalence.
However, unlike training a deep neural network by minimizing the classification error, training a deep neural network for metric learning is notoriously more difficult~(\cite{qian2018large,wang2017deep}).  Many studies have attempted to address this challenge by focusing on several issues. The first issue is how to define a loss function over pairs of examples. A variety of  loss functions have been proposed  such as contrastive loss~(\cite{hadsell2006dimensionality}),  binomial deviance loss~(\cite{yideep}),  margin loss~(\cite{wu2017sampling}), lifted-structure (LS) loss~(\cite{oh2016deep}), N-pair loss~(\cite{sohn2016improved}), triplet loss~(\cite{schroff2015facenet}), multi-similarity (MS) loss~(\cite{wang2019multi}.
%These above loss functions are known as \emph{pair-based} loss. 
%All of them follow the philosophy that preserves the distance between similar data points close and dissimilar data points far from the embedding space. 
The major difference between these pair-based losses lies at  how the pairs interact  with each other in a mini-batch. In simple pairwise loss such as binomial deviance loss, contrastive loss, and margin loss, pairs are regarded as independent of each other.  In triplet loss, a positive pair only interacts with one negative pair. In N-pair loss, a positive pair interacts with all negative pairs. In LS loss and MS loss, a positive pair interacts with all positive pairs and all negative pairs.  The trend is that the loss functions become increasingly complicated but are difficult to understand. 
 
%-----------------------------------------
%The major difference between these pair-based losses lies at  how the pairs interact  with each other in a mini-batch. In simple pairwise loss such as binomial deviance loss, contrastive loss, and margin loss, pairs are regarded as independent of each other. 
%In triplet loss, a positive pair only interacts with one negative pair. In N-pair loss, a positive pair interacts with all negative pairs. In LS loss and MS loss, a positive pair interacts with all positive pairs and all negative pairs.
%-----------------------------------------

%For pair-based deep metric learning,

In parallel with the loss function, how to select informative pairs to construct the loss function has also received great attention. Traditional approaches  that construct pairs or triplets over all examples before training suffer from prohibitive  $O(N^2)$ or $O(N^3)$ sample complexity, where $N$ is the total number of examples. To tackle this issue, constructing pairs within a mini-batch is widely used in practice. Although it helps mitigate the computational and storage burden, %the number of pairs in a batch are polynomial growth as the number of batch size increases, which would cause information redundancy. 
slow convergence and model degeneration with inferior performance still commonly exist when using all pairs in a mini-batch to update the model.  To combat this issue, various  {\it pair mining} methods have been proposed to complement the design of loss function, such as hard (semi-hard) mining for triplet loss~(\cite{schroff2015facenet}), distance weighted sampling (DWS) for margin loss~(\cite{wu2017sampling}), MS sampling for MS loss~(\cite{wang2019multi}). These sampling methods usually keep all  positive (similar) pairs and select roughly the same order of negative (dissimilar) pairs according to some criterion.%, such as over-sampling and under-sampling.

%There are two different ways to construct pairs or triplets during the training processes. 
% The first one is to construct pairs or triplets based on all samples at the beginning of each epoch, which is denoted as offline pair construction. 
% The other method is to construct pairs within a mini-batch of samples in every iteration, which is called online pair construction.
% The offline method usually suffers from $O(N^3)$ sample complexity that leads to unacceptable computation time and storage requirements.
% However, online pair construction is much more efficient since it only considers the data in a batch, and thus widely used in practice. 

% Even though constructing pairs within a batch alleviates the information redundancy problem caused by uniformly sampling from the entire dataset, the number of pairs in a batch are still polynomial growth with increasing number of batch size. 
% Slow convergence and model degeneration with inferior performance would exist when using all pairs in a batch to learn an embedding. 
% Thus, recently proposed loss functions are often accompanied with {\it pair mining} methods to select the most informative pairs, such as hard (semi-hard) mining for triplet loss~(\cite{schroff2015facenet}), distance weighted sampling(DWS) for margin loss, MS sampling for MS loss~(\cite{wang2019multi}).

Regardless of these great efforts, existing studies either fail to explain the most fundamental problem in DML or fail to propose most effective approach towards addressing the fundamental challenge. It is evident that the loss functions become more and more complicated. But it is still unclear why these complicated losses are effective and how does the pair mining methods affect the overall loss within a mini-batch. 
%have overlooked the fundamental root cause of the inferior performance in DML, i.e., the sheer imbalance between positive pairs and negative pairs. Only some naive solutions are adopted in recently proposed methods to keep the number of positive pair and negative pair roughly the same, such as over-sampling and under-sampling. Unfortunately, these methods may not be often effective in real tasks. Since these selected pairs determines the derivatives for updating the model, it is urgent to find a solution to handle the pair imbalance problem.
In this paper, we propose a novel effective solution to DML and bring new insights from the perspective of learning theory that can guide the discovery of new methods. Our philosophy  is simple: casting the problem of DML into a simple pairwise classification problem and focusing on addressing the most critical issue, i.e., the sheer imbalance between positive pairs and negative pairs. To this end, we employ simple pairwise loss functions (e.g., margin loss, binomial deviance loss) and propose a flexible  distributionally robust optimization (DRO) framework for defining a robust loss over pairs within a mini-batch. The idea of DRO is to assign different weights to different pairs that are optimized by maximizing the weighted loss over an uncertainty set for the distributional variable. The model is updated by stochastic gradient descent  with stochastic gradients computed based on the sampled pairs according to the found optimal  distributional variable.

The DRO framework allows us to (i) connect to  advanced learning theories that already exhibit their power for imbalanced data, hence providing theoretical explanation for the proposed framework; (ii)  to unify pair sampling and loss-based methods to provide a unified perspective for existing solutions; (iii) to induce simple and effective methods for DML, leading to state-of-the-art  performance on several benchmark datasets.
 %that samples pairs particularly to promote the performance for DML.Compared to existing methods, our method is very general and allows users to apply new sampling criteria by defining different formulations.
 %Additionally, when reducing our framework to these two special cases, we demonstrate that our sampling framework automatically perform pair mining, while pair mining is usually a separated procedure in other methods.
%Meanwhile, our framework explains the reason why MS loss and LS loss functions can learn a better embedding -- approximating a good pair distribution by interacting with different pairs.
%Furthermore, we hope that our framework offers a new insight to design new solutions to pair imbalance in DML.
The contributions of our work are summarized as follows:
\begin{itemize}[leftmargin=*]
\vspace*{-0.1in}
    \item We  propose a general solution  framework for DML, i.e., by defining a robust overall loss based on the DRO formulation and updating the model based on pairs sampled according to the optimized sampling probabilities. We provide theoretical justification of the proposed framework from the perspective of advanced learning theories. 
   % formulate the sampling process of DML into a unified framework which is implemented based on DRO.
    %We design three different formulations of our DRO framework to handle pair imbalance problem. 
    
    \item We show that the general DRO framework can recover existing methods based on complicated pair-based losses: LS loss and MS loss by specifying different uncertainty sets for the distributional variable in DRO.
    It verifies that our method is general and brings a unified perspective regarding pair sampling and complicated loss over all pairs within a batch. 
   % Specifically, we show the recently proposed complicated  losses, i.e., LS and MS loss functions, can be included as special cases of our framework by specifying different uncertainty set for the distributional variable, which encodes some prior knowledge for regularization. 

    \item We also propose simple solutions under the general DRO framework for tackling DML. Experimental results show that our proposed variants of DRO framework outperform state-of-the-art methods on several benchmark datasets.%, including the designed loss functions and the sampling methods.
\end{itemize}

\vspace*{-0.1in}
\section{Related Work}
\vspace*{-0.1in}
\textbf{Loss Design.}
%As other tasks, loss is crucial to learn a good embedding in DML.
The loss function is usually defined over the similarities or distances between the induced feature embeddings of pairs. There are simple pairwise losses that simply regard DML as binary classification problem using averaged loss over pairs, e.g., contrastive loss, binomial loss, margin loss.  It is notable that the binomial loss proposed in~(\cite{yideep}) assigns asymmetric weights for positive and negative pairs, which can mitigate the issue of imbalance to certain degree. % can be viewed as making use of a pre-defined cost-sensitive learning information for both positive and negative pairs.
The principal of the newly designed  complicated \emph{pair-based} losses can be summarized as heuristically discovering specific kinds of relevant information between groups of pairs to boost the training. The key difference between these complicated losses lies at how to group the pairs. N-pair loss put one positive pair and all negative pairs together, Lifted-structure loss and MS-loss group all positive pairs together and all negative pairs together for each example. 
%or instance, binomial deviance assigns asymmetric weights to positive pairs and negative pairs respectively to address the issue that the negative pairs
%prone to under-fitting~(\cite{yideep}). 
%From DRO perspective, binomial loss can be interpreted as assign a constant $p_+$ to all positive pairs and $p_-$ to all negative pairs in terms of gradient updates to handle the pair imbalance problem.
%For instance, N-pair loss is defined upon the summation of the relative similarity between $N-1$ negative pairs and one positive pair of each anchor (\cite{sohn2016improved}).
%Lifted-structure loss takes the lifted structure of the embedding between all positive pairs and negative pairs of each anchor into consideration~(\cite{wang2017deep}).
%The asymmetric weights for positive and negative pairs designed in binomial loss can be viewed as making use of a pre-defined cost-sensitive learning information for both positive and negative pairs. 
%Thus, 
%MS loss, a carefully combination of binomial loss and lifted structure loss, can be viewed as making use of two different types of relevant information between pairs. 
In contrast, our DRO framework employs simple pairwise loss but induce complicated overall loss in a systematic and interpretable way. 
%Our DRO framework not only unifies these types of relevant information into one framework by setting different \emph{uncertainty decision set}, but also is able to provide theoretical explanations for them.

% loss  aims to  by optimizing an $(N+1)$-tuplet objective function which is constructed by the positive pair and its corresponding $(N-1)$-negative pairs carefully selected from $N-1$ different classes(\cite{sohn2016improved}}), while lifted-structure loss is developed according to the lifted structure embedding over all positive and negative pairs of samples~(\cite{wang2017deep}). The relative information between different 

%It is worth to mention that which can also be interpreted as a method to handling the pair imbalance problem from the perspective of cost sensitive learning.Then the relative information between positive and negative pairs are taken into consideration when updating gradients

\textbf{Pair Mining/Pair Weighting.}
~\cite{wu2017sampling} points out that pair mining plays an important role in distance metric learning. Different pair mining methods have been proposed, including semi-hard sampling for triplet loss, distance weighted sampling (DWS) for margin loss, MS mining for MS losses. These pair mining methods aim to select the hard positive and negative pairs for each anchor. 
For instance, \cite{schroff2015facenet} selects the hard negative pairs whose distance  is smaller than that between the positive pairs in  triplets,~\cite{shi2016embedding} selects the hardest positive pair whose distance is smaller than that of the nearest negative pair in a batch, and MS mining~(\cite{wang2019multi}) selects hard negative pairs whose distance is smaller than the largest distance between positive pairs and hard positive pairs whose distance is larger than the smallest distance between negative pairs at the same time.
 DWS method keeps all positive pairs but samples negative pairs according to their distance distribution within a batch. %To address the issue that too much hardest negative pairs will impede the convergence of model at early stage, SemiHard (SH) is used in the experiments of \cite{schroff2015facenet} for triplet loss. 
The proposed DRO framework induce a pair sampling method by using the optimal distributional variables that defines the robust loss over pairs within a mini-batch. %we proposed can be view as a pair mining methods that select the hard negative and hard positive pairs in terms of large losses. 
As a result, the sampling probabilities induced by our DRO framework is automatically adaptive to the \emph{pair-based losses}. There are other works that study the problem from the perspective of pair weighting instead of pair sampling.  
For example, \cite{yu2018hard} heuristically design exponential weights for the different pairs in a triplet, which is a special case of our DRO framework. Details are provided in the supplementary. However, since the quality of anchors varies very much, it may not be reasonable to sample the same number of pairs from all anchors.
%\textbf{General Pair Weighting.}
%\cite{wang2019multi} propose a general framework to explain different \emph{pair-based losses} can be viewed as different weighting scheme to the pairs in terms of similarity, and develop a new MS loss by combing the weighting scheme of binomial loss and lifted-structure loss. Similar weighting idea has also been applied in
%For example, \cite{yu2018hard} heuristically design exponential weights and polynomial weights for the different pairs in a triplet. \textcolor{red}{In the supplementary, we showed that assigning exponential weights to each pair is a special case of our DRO framework}.

% Instead, we should emphasize more on other samples. Furthermore, all of above weighting scheme only a special case of our DRO framework.

\textbf{Imbalance Data Classification.} There are many studies in machine learning which have tackled the imbalanced issue. Commonly used tricks include over-sampling,  under-sampling and cost-sensitive learning. However, these approaches do not take the differences between examples into account.  Other effective approaches grounded on advanced learning theories include minimizing maximal losses~\citep{shalev2016minimizing},  minimizing top-k losses~\citep{fan2017learning} and minimizing variance-regularized losses~\citep{namkoongnips2017variance}. However, these approaches are not efficient  for deep learning with big data, which is a severe issue  in DML.  In contrast, the proposed DRO formulation is defined over a mini-batch of examples, which inherits the theoretical explanation from the literature and is much more efficient for DML. In addition, the induced loss by our DRO formulation include maximal loss, top-k loss and variance-regularized loss as special cases by specifying different uncertainty sets of the distributional variable.

\section{DML As A DRO-Based Binary Classification Problem}
\vspace*{-0.1in}\label{sec:KeyTechnique}
In this section, we will first present a general framework for DML based on DRO with theoretical justification. We will then discuss three simple variants of the proposed framework and also show how the proposed framework recover existing complicated losses for DML. 

{\bf Preliminaries.} Let $\x\in\mathbb{R}^D$ denote an input data (e.g., image) and $f(\cdot;\theta):\mathbb{R}^D\rightarrow \mathbb R^d$ denote the feature embedding function defined by a deep neural network parameterized by $\theta$. The central task in DML is to update the model parameter $\theta$ by leveraging  pairs of similar and dissimilar examples. Following most existing works, at each iteration we will sample a mini-batch of examples denoted by $\{ \x_1, ..., \x_B \}$. We can construct $B^2$ pairs between these examples~\footnote{For simplicity, we consider all pairs including self-pair.}, and let $y_{ij}$ denote the label of pairs, i.e., $y_{ij} = 1$ if the pair is similar (positive),  and $y_{ij} = 0$ if the pair is dissimilar (negative). The label of pairs can be either defined by users or derived from the class label of individual examples.  Existing works of DML follow the same paradigm for learning the deep neural network i.e., a loss function $F(\theta)$ is first defined over the pairs within a mini-batch and the model parameter $\theta$ is updated by gradient-based methods. Various gradient-based methods can be used, including stochastic gradient descent (SGD), stochastic momentum methods and adaptive gradient methods (e.g. Adam). Taking SGD as an example, the model parameter $\theta$ can be updated by
$\theta \leftarrow \theta - \eta \nabla F(\theta)$, 
where $\eta$ denotes the learning rate. The focus here is to how to define the loss function $F(\theta)$ over all pairs within a mini-batch.  As mentioned earlier, we will cast the problem as simple binary classification problem, i.e., classifying a pair into positive or negative. To this end,  we use $l_{ij}(\theta)=l(f( \x_i;\theta), f(\x_j;\theta), y_{ij})$ denote the pairwise classification loss between $\x_i$ and $\x_j$ in the embedding space (e.g., margin loss~\cite{wu2017sampling}, binomial loss~\cite{yideep}).  A naive approach for DML is to use the averaged loss over all pairs, i.e., \useshortskip
%\begin{align*}
$F_{\text{avg}}(\theta) = \frac{1}{B^2}\sum_{i=1}^{B}\sum_{j=1}^{B}l_{ij}(\theta)$. 
%\end{align*}
However, this approach will suffer from the severe imbalanced issue, i.e., most pairs are negative pairs. The gradient of $F_{\text{avg}}$ will be dominated by that of negative pairs. 
%Let $[B]$ denote $\{ 1, 2, ..., B\}$.
%how we construct a general DRO formulation for pair sampling in a batch of data for DML.
% Then we verify the generality of our DRO framework by showing that the recent MS and LS loss can be recovered by our method and thus included as special cases. 
% Further we design three new constrain sets to from the perspective of pair imbalance problem within a batch.
%Existing DML methods usually leverage all pairs in a batch or only apply naive approaches to derive balanced positive and negative pairs. As stated by many previous studies \cite{ge2018deep,wang2019multi,wu2017sampling}, there can be much information redundancy in these pairs. Thus, the subgradient computed by these pairs may not be sufficiently informative. To handle this issue, we propose to cast DML as a pairwise binary classification problem and formulate it to a DRO-based framework.

\subsection{General DRO-Based Framework}
\vspace*{-0.1in}%We define the similarity of two samples as $S_{ij} := \langle f(\x_i; \theta), f(x_j;\theta) \rangle$, where $\langle \cdot,\cdot \rangle$ denotes dot product. This results in a similarity matrix ${\bf S}$ with size of $B \times B$.
To address the imbalanced pair issue, we propose a general DRO formulation to compute a robust loss.  The formulation of our DRO-based loss over all pairs within a mini-batch is given below: \useshortskip
\begin{align}
\label{eq:general_DRO_framework}
 F(\theta) = \max_{\p \in \mathcal U}\{ g(\theta, \p) : = \sum_{i=1}^B \sum_{j = 1}^B p_{ij} l_{ij} (\theta)\},
\end{align}
where $\p \in \mathbb R_+^{B^2}$ is a non-negative vector with each element $p_{ij}$ representing  a weight for an individual pair. $\mathcal U \subseteq \mathbb R^{B^2}$ denotes the decision set of $\p$, which encodes some prior knowledge about $\p$. In the literature of DRO~\cite{namkoongnips2017variance}, $\p$ is interpreted as a probability vector such that $\sum_{ij}p_{ij}=1$ called the distributional variable and $\mathcal U$ denotes the uncertainty set that specifies how $\p$ deviates from the uniform probabilities $(1/B^2, \ldots, 1/B^2)$. In next subsection, we will propose simple variants of the above general framework by specifying different constraints or regularizations for $\p$. Below, we will provide some theoretical evidences to justify the above framework.

%For simplicity, we let $g(\theta, \p) = \sum_{i=1}^B \sum_{j = 1}^B p_{ij} l_{ij} (\theta)$ be the objective function.
%The key of the above framework is to flexibly define the objective function $g(\theta, \p)$ and uncertainty decision set $\mathcal U$ for $\p$.
%$F(\theta)$ is thus a robust loss of $\theta$.
%Then our framework works as follows.
%Given $\theta$, we first find the optimal solution of $\p$, denoted by $\p^*$.
%Then we compute the subgradient of $g$ w.r.t. $\theta$, i.e., $\partial_{\theta} g(\theta, \p^*)$, for updating the model.

% However, when $F(\theta)$ is non-convex, $\partial F(\theta)$ is not directly accessible due to the inner subproblem.
% Our framework provides an alternative way to compute the subgradient of $\theta$.
% Given $\theta$, we first find the optimal solution of $\p$, denoted by $\p^*$.
% Then we compute the subgradient of $g$ w.r.t. $\theta$, i.e., $\partial_{\theta} g(\theta, \p^*)$, for updating the model.
% Even if $\partial_{\theta} g(\theta, \p^*)$ is not necessary equal to  $\partial F(\theta)$, there is an immediate connection between them, as shown in Lemma \ref{lemma:argmax_is_grad} later.

To theoretically justify the above loss, we connect~(\ref{eq:general_DRO_framework}) to exiting works in machine learning by considering three different uncertainty sets for $\p$.  First, we can consider a simple constraint $\mathcal U =\Delta= \{p_{ij}\geq 0, \sum_{ij}p_{ij}=1\}$. As a result, $F(\theta) = \max_{ij}l_{ij}(\theta)$ becomes the maximal loss over all pairs. \cite{shalev2016minimizing} shows that minimizing maximum loss is robust to imbalanced data distributions and also derives better generalization error  for imbalanced data with a rare class. However, the maximal loss is more sensitive to outliers~\citep{zhu2019robust}. To address this issue, top-$K$ loss~\citep{fan2017learning} and variance-regularized loss~\citep{namkoongnips2017variance} are proposed, which can be induced by the above DRO framework.  If we set  $\mathcal U = \{\sum_{ij}p_{ij}=1, 0\leq p_{ij}\leq 1/K\}$, $F$ will become the top-$K$ loss $F(\theta)=\frac{1}{K}\sum_{i=1}^K l_{[i]}(\theta)$, where $l_{[i]}(\theta)$ denotes the $i$-th largest loss over all pairs. If we set $\mathcal U_\phi = \{\p\in\Delta, D_\phi(\p\| \mathbf 1/B^{2})\leq \frac{\rho}{B^2}\}$, where $D_\phi (\p \| \p') = \int \phi(\frac{d\p}{d\p'}) d\p'$ is the $\phi$-divergence between two distributions $\p$ and $\p'$  with  $\phi(t) =\frac{1}{2}(t- 1)^2$, then the DRO-based loss becomes the variance-regularized loss under certain condition about the variance of the random loss, i.e., for a set of i.i.d random losses $\{\ell_1, ..., \ell_n \}(n=B^2)$ we could have \useshortskip
\begin{align*}
\sup_{\p\in\mathcal U_\phi} \sum_{i=1}^n p_i \ell_i  = \frac{1}{n}\sum_{i=1}^n \ell_i + \sqrt{ \frac{ 2\rho \text{Var}_{n} (\ell) }{ n } },
\end{align*} where $\text{Var}_{n} (\ell) $ denotes the empirical variance of $\ell_1, ..., \ell_n $. We can see that the second term in R.H.S of the above equation  involves the variance, which  can play a role of regularization.  The variance-regularized loss has been justified from advanced learning theory by~\cite{namkoongnips2017variance}, and its promising performance for imbalanced data has been observed as well.  %as indicated by  the following theorem. ore diving into the specific variants of our framework in the next subsection,

Before ending this subsection, we will discuss how to update the model parameter $\theta$ based on  the robust loss $F(\theta)$ defined by~(\ref{eq:general_DRO_framework}). A simple approach is to find an optimal distributional variable $\p_*$ to (\ref{eq:general_DRO_framework}) and then update $\theta$ according to  the subgradient of weighted loss by  $\partial_\theta g(\theta, \p^*) = \sum_{ij}p^*_{ij}\nabla l_{ij}(\theta)$, which is justified by the following lemma. %the maximization problem of $\p$  %Indeed, there is a gap between $\partial F(\theta)$ and $\partial_\theta g(\theta, \p)$: $\partial_{\theta} g(\theta, \p^*)$ is not necessary equal to $\partial F(\theta)$ when $F(\theta)$ is non-convex in $\theta$.
%However, as shown in Lemma \ref{lemma:argmax_is_grad} later, there is an immediate connection between them, and we can precisely compute .
%Thirdly, the subgradient $\partial_\theta g(\theta, \p^*)$, which we use to update the model as mentioned, has an immediate connection to $\partial F(\theta)$.
%% Since $\partial F(\theta)$ is not directly accessible, our framework computes $\partial_\theta g(\theta, \p^*)$ instead.
%The following lemma states how $\partial_\theta g(\theta, \p^*)$ relates to $\partial F(\theta)$.
\vspace*{-0.1in}\begin{lem}\label{lemma:argmax_is_grad}
% (Gradient of robust loss in $\theta$ is contained by $\partial_\theta g(\theta, \p^*(\theta))$, Theorem 10.13 in \cite{rockafellar2009variational})
Assume that $g$ is proper, lower-semicontinuous  in $\theta$ and level-bounded in $\p$ locally uniformly in $\theta$.
Then the subgradient $\partial F(\theta) \subset \bigcup_{\p^* \in P^*(\theta)} \partial_{\theta} g(\theta, \p^*)$, where $P^*(\theta)$ denotes the optimal solution set of the maximization problem in (\ref{eq:general_DRO_framework}). 
Furthermore, when $l_{ij}(\theta)$ is smooth in $\theta$ and $P^*(\theta)$ is a singleton, i.e., $\p^* = \arg\max_{p} g(\theta, \p)$ is unique,
we have $\partial F(\theta) = \partial_\theta g(\theta, p^*)$.
\end{lem}
\vspace*{-0.1in}\begin{remark}
The above lemma can be proved by Theorem 10.13 in \cite{rockafellar2009variational}.
It shows that even if we may not directly compute $\partial F(\theta)$, our framework can at least obtain its superset $\partial_\theta g(\theta, \p^*)$.
Particularly, if we have additional conditions, i.e., $l_{ij}(\theta)$ is smooth in $\theta$ and the optimal solution $\p^*$ is unique (considering our regularized formulation below), it theoretically guarantees that our framework exactly computes $\partial F(\theta)$.
\end{remark}

\subsection{Proposed Three Variants of Our Framework}
\vspace*{-0.1in}In this subsection, we present three variants of our general framework.
%Suppose that our goal is to sample pairs in a batch for computing the subgradient of the model to promote training.
%The first two variants directly select $K$ pairs with largest pairwise loss over all pairs in a batch, while the last one maintains a distribution over all pairs and samples pairs according this distribution.
In order to contrast to other variants recovering existing complicated losses presented in next subsection, we introduce some notations and make some simplifications. 
For each example $\x_i$ that serves  as an anchor data, let $\P_i =\{j|y_{ij}=1 ,j\in [B]\}$ and $\N_i = \{j| y_{ij}=0, j\in [B]\}$ denote the index sets of positive and negative pairs, respectively.
%In addition, we denote by $\P_i^+ =\{j|y_{ij}=1 , l_{ij}(\theta) > 0, j\in [B]\}$ and $\N_i^+ = \{j|y_{ij}=0,l_{ij}(\theta) > 0, j\in [B]\}$ the index sets of those pairs with non-zero loss.
%It is easy to see that $\P_i^+ \subseteq \P_i$, and $\N_i^+ \subseteq \N_i$.
Let $\P = \bigcup_{i=1}^B \P_i$ and $N = \bigcup_{i=1}^B \N_i$.
We denote the cardinality of a set by $P = |\P|$.
For simplicity, we let $P_i = |\P_i|$, $N_i = |\N_i|$, $P = |\P|$ and $N = |\N|$. Since  zero losses usually do not contribute to the computation of the subgradient for updating the model, we can simply eliminate those examples for consideration.

The first variant is to simply select the top-$K$ pairs with $K$-largest losses, which is equivalent to the following DRO formulation: \useshortskip
\begin{align*}%\label{equ:DRO-TopK-A}
& \text{DRO-TopK: }
\max_{ \p }  
% ~&~ 
\sum\limits_{i=1}^{B} \sum\limits_{j\in \P_i\cup \N_i} p_{ij} l_{ij}(\theta) 
,~ \text{s.t.} ~
\sum\limits_{i=1}^{B}\sum\limits_{j\in \P_i\cup \N_i} p_{ij}= 1, 0\leq p_{ij}\leq 1/K   ,
%\\
%\label{equ:DRO-TopK-PN}
%& \text{DRO-TopK-PN: }
%\max_{\p \in \{ 0, 1 \}^{P^+ + N^+}}
%% ~&~
%\sum\limits_{i=1}^{B} \sum_{j \in \P_i^+\cup \N_i^+} p_{ij} l_{ij} (\theta) 
%~ \text{s.t.}~
%\sum_{i=1}^{B} \sum_{j \in \P_i^+} p_{ij} \leq \frac{K}{2}, 
%\sum_{i=1}^{B} \sum_{j \in \N_i^+} p_{ij} \leq \frac{K}{2}   .
\end{align*}
where $K$ is a hyper-parameter. The gradient of the robust loss can be simply computed by sorting the pairwise losses and computing the average of top-$K$ losses.  

The second variant is a variant of the variance-regularized loss. Instead of specifying the uncertainty set $\mathcal U_{\phi}$, we use a regularization term for the ease of computation, which is defined by\useshortskip
\begin{equation*}
\begin{aligned}%\label{equ:DRO-TopK-PS}
& \text{DRO-KL:} 
\max_{\p  \in  \mathbb R_+^{ P + N }  }  
\sum_{i=1}^{B} \sum_{ j \in \P_i \cup \N_i } p_{ij} l_{ij} (\theta)
- \gamma D_{KL}(\p||\frac{ {\bf 1} }{ P + N } ), 
~ \text{s.t.}~
\sum_{i=1}^{B} \sum_{ j \in \P_i \cup \N_i } p_{ij} = 1 ,
\end{aligned}
\end{equation*}
where $\gamma>0$ is a hyper-parameter and $D_{KL}$ denotes the KL divergence between two probabilities. The optimal solution to $\p$ can be easily computed following~\cite{namkoong2016stochastic}.
It is notable that the optimal solution $\p^*$ is not necessarily sparse. Hence, computing $\sum_{ij}p^*_{ij}\nabla l_{ij}(\theta)$ needs to compute the gradient of pairwise loss for all pairs, which could be prohibitive in practice when the mini-batch size is large. To alleviate this issue, we can simply sample a subset of pairs according to probabilities in $\p^*$ and the compute the averaged gradient of these sampled pairs. 
%The KL divergence regularizer of $\p$ in the objective function is originally an element of the uncertainty decision set $\mathcal{U}$ of $\p$.
%We move it to the objective function as a regularization term with a hyperparameter $\gamma$ to avoid additional overhead handling the constraint.
%The remaining part of the constraints ensure $\p$ is in a probability simplex.
% The greatest benefit of this variant is the flexibility and generality.
% In the next subsection, we show how we can derive SOTA pair-based loss functions such as LS and MS loss.
% This variant serves as an inspiration to 1) explain why SOTA loss functions performs well; 2) cast light on new formulations.

The third variant of our DRO framework is explicitly balancing the number of positive pairs and negative pairs by choosing top $K/2$ pairs for each class, which is denoted by DRO-TopK-PN: \useshortskip
\begin{align*}%\label{equ:DRO-TopK-A}
%\label{equ:DRO-TopK-PN}
& \text{DRO-TopK-PN: }
\max_{\p \in \{ 0, 1 \}^{P + N}}
% ~&~
\sum\limits_{i=1}^{B} \sum_{j \in \P_i\cup \N_i} p_{ij} l_{ij} (\theta) 
, ~ \text{s.t.}~
\sum_{i=1}^{B} \sum_{j \in \P_i} p_{ij} \leq \frac{K}{2}, 
\sum_{i=1}^{B} \sum_{j \in \N_i} p_{ij} \leq \frac{K}{2}   .
\end{align*}
%The key difference between DRO-TopK-PN and DRO-TopK is whether enforcing the numbers of selected positive and negative pairs equal to each other.
%Consequently, DRO-TopK selects $K$ pairs with largest loss, while 
For implementation, we can simply select $K/2$ positive pairs with largest losses and $K/2$ negative pairs with largest loss respectively, and compute averaged gradient of the pairwise losses of the selected pairs for updating the model parameter. 
%DRO-TopK-PN ensures balanced numbers of positive and negative pairs.

%We would emphasize that our framework is general, and other new formulations can be included by the DRO-based framework. For example, in the next subsection, we showcase that with slight modification on the variant DRO-KL, our framework recovers SOTA pair-based losses.

\subsection{Recovering the Method based on SOTA Pair-Based Loss}
\vspace*{-0.1in}\label{section:recover_LS_loss}
Next we show that proposed framework can recover the method based on SOTA complicated losses.  With the induced  feature vector $f(\x; \theta)$ normalized to have unit norm, we define the similarity of two samples as $S_{ij} := \langle f(\x_i; \theta), f(\x_j;\theta) \rangle$, where $\langle \cdot,\cdot \rangle$ denotes dot product.  Specifically, we consider two SOTA loss functions, LS and MS loss, which are defined below:\useshortskip% are special cases of the proposed DRO-KL variant in (\ref{equ:DRO-TopK-PS}) with different uncertainty decision set $\mathcal{U}$.
%MS Loss and LS loss are defined as follows
\begin{align}
&
\mathcal L_{MS} = \frac{1}{n}\sum_{i=1}^{n}\{\frac{1}{\alpha}\log[1 + \sum_{k\in \P_i}e^{-\alpha(S_{ik}-\lambda)}] + \frac{1}{\beta}\log[1 + \sum_{k\in \N_i}e^{\beta(S_{ik}-\lambda)}]\}
\label{equ:LS}
\\
&
\mathcal L_{LS} = \sum_{i=1}^{n} [\log\sum_{k\in\P_i}e^{\lambda - S_{ik}} + \log\sum_{k\in \N_i}e^{S_{ik}-\lambda}]_+ .
\label{equ:MS}
\end{align}
where $\alpha, \beta, \lambda$ are hyper-parameters of these losses. 

The key to our argument is that the gradient computed based on these losses can be exactly computed according to our DRO framework by choosing appropriate constrained set $\mathcal U$ and setting the pairwise loss as the margin loss. To this end, we first show the gradient based on the LS loss, which  can be computed by~\citep{wang2019multi}:\useshortskip
%It can be formulated as a function in terms of the similarity matrix $S$ and $y$: $\mathcal{L}(S, y)$.
%The derivative w.r.t. $\theta$ can be computed by
\begin{align}
\frac{ \partial \mathcal L(S) }{ \partial \theta }
=
\frac{ \partial \mathcal L(S) }{ \partial S } \cdot \frac{ \partial S }{ \partial \theta }
= \sum_{i=1}^B \sum_{j=1}^B \frac{ \partial \mathcal L(S) }{ \partial S_{ij} } \cdot \frac{ \partial S_{ij} }{ \partial \theta } 
\end{align}\useshortskip
which can be alternatively written as 
\useshortskip
\begin{equation}
\label{equ:Closed-form-Weighting}
    \begin{aligned}
\frac{\partial \mathcal L(S) }{\partial \theta}   = \sum\limits_{i=1}^B  \Big(\sum\limits_{j\in\mathcal N_i} w^-_{ij}\frac{\partial S_{ij}}{\partial \theta} -\sum\limits_{j\in\P_i} w^+_{ij}\frac{S_{ij}}{\partial \theta} \Big ). 
    \end{aligned}
\end{equation}
It can be shown that for LS loss, derivations are provided in the supplementary, we have\useshortskip 
\begin{equation}
\label{equ:w-ij+LS}
\begin{aligned}
w_{ij}^+ = \frac{e^{\lambda-S_{ij}}}{\sum\limits_{k\in \P_i}^B e^{\lambda-S_{ik}}} 
=  \frac{1}{\sum\limits_{k\in\P_i}^B e^{S_{ij}-S_{ik}}} , \quad w_{ij}^- = \frac{e^{S_{ij} - \lambda}}{\sum\limits_{k\in \N_i}^B e^{S_{ik}-\lambda}} =  \frac{1}{\sum\limits_{k\in \N_i}^B e^{S_{ik} -S_{ij}}}     .
\end{aligned}
\end{equation}
%Then the weight for a negative pair $\{x_i, x_j\}$ is computed as
%\begin{equation}
%\label{equ:w-ij-LS}
% \begin{aligned}
%w_{ij}^- = \frac{e^{S_{ij}}}{\sum\limits_{y_k = y_i}^B e^{\lambda-S_{ik}}} =  \frac{1}{\sum\limits_{y_k = y_i}^B e^{S_{ik} -S_{ij}}}     .
%\end{aligned}
%\end{equation}
%where $w_{ij} = | \frac{\partial \mathcal{L}(S, y)}{S_{ij}} |$, which is the weight for $S_{ij}$. 
%For simplicity, we denote $w_{ij}^+ = w_{ij}$ when $y_i = y_j$, and $w_{ij}^- = w_{ij}$ when $y_i \neq y_j$.
%Next, we show the the representation for LS loss and MS loss under GPW, respectively.
To recover the gradient of the LS loss under our DRO framework, we employ the pairwise  margin loss for $l_{ij}(\theta)$, i.e.,  $l_{ij}(\theta) = [ m + y_{ij} (\lambda - S_{ij}) ]_{+}$,
where $m$ and $\lambda$ are two hyper-parameters and $[a]_+ = \max\{ 0, a \}$. Assume that the margin parameter $m$ is sufficiently large such that $l_{ij}(\theta)>0$ for all pairs. 
%Then we show that the optimal solution $\p^*$ is identical to the weight of the LS loss under the general pair weighting (GPW) in \cite{wang2019multi}.
The key to deriving the same gradient of the LS loss under our framework is to group distributional variables in $\p$ for the positive and negative pairs according to the anchor data.  Let $\p_i^+ \in \mathbb R^{P_i}$ and $\p_i^- \in \mathbb R^{N_i}$ denote the corresponding  vectors of positive and negative pairs for the anchor $\x_i$, respectively. 
%Then $\p \in \mathbb R^{P^+ + N^+}$ is constructed by concatenating all $\p_i^+$ and $\p_i^-$ for $i\in [B]$ together.
%Specifically, compared to DRO-KL in (\ref{equ:DRO-TopK-PS}), we apply different KL constraints and simplex constraints on $\p$.
%Instead of using only one KL constraint and one simplex constraint on the entire $\p$, we leverage grouped constraints on each $\p_i^+$ and $\p_i^-$, for $i\in [B]$ as follows
%Let us first consider recovering the LS loss.
Let us  consider the following DRO formulation:\useshortskip  
\begin{equation}
\begin{aligned}\label{equ:DRO_Lifted_Structure_Loss}
F(\theta) = \max_{\p \in \mathbb R_+^{ P + N } } & \sum_{i=1}^{B} \sum_{j \in \P_i \cup \N_i} p_{ij} l_{ij}(\theta) 
- \sum\limits_{i=1}^{B} \Big( \gamma_i^+ D_{KL}(\p_i^+ || \frac{\textbf{1}}{ P_i }) + \gamma_i^- D_{KL}(\p_i^- || \frac{\textbf{1}}{ N_i }) \Big) \\
\text{s.t.}  
~&~
\sum_{j \in \P_i} p_{ij} = 1, 
\sum_{k \in \N_i} p_{ik} = 1,~ \text{for } i \in [B]   ,
\end{aligned}
\end{equation}
where $\gamma_i^+ \geq 0$ and $\gamma_i^- \geq 0$ for $i\in[B]$ are hyper-parameters. %Since the objective value is convex in terms of $\p_i^+, \p_i^-, i \in [B]$, we are able to derive the closed form of the optimal $\p^*$ by making use of KKT conditions.
we can easily derive the closed-form solution for $\p_*$, i.e.,  %Here we only provide the closed form in equation as follows
%\begin{align*}
$ p_{ij}^{+*} 
= \frac{1}{\sum\limits_{k \in \P_i}e^{ ( S_{ij}-S_{ik} ) / \gamma_i^+ } }$, and $p_{ij}^{-*} 
= \frac{1}{\sum\limits_{k\in \N_i}e^{ ( S_{ik}-S_{ij} ) / \gamma_i^- } } $.   
%\end{align*}
Then computing the gradient of the robust loss with respect to $\theta$ by using the above optimal $\p^*$, we have: \useshortskip
\begin{align*}
\partial F(\theta) = \sum_{i=1}^{B} \Big(\sum_{j \in \N_i} p_{ij}^{-*} \frac{\partial S_{ij}}{\partial \theta} - \sum_{j \in \P_i} p_{ij}^{+*} \frac{\partial S_{ij}}{\partial \theta} \Big) 
\end{align*} 
which exactly recover the gradient in~(\ref{equ:w-ij+LS}) by setting  $\gamma^+_i = \gamma^-_i=1$. 

%On the other hand, in \cite{wang2019multi}, LS loss and MS loss are included in their proposed general pair weighting (GPW) framework.
%In GPW, for the similarity matrix, each element $S_{ij}$ is weighted by a weight $w_{ij}$.
%Then a loss function can be formulated as a weighted combination of all elements of the similarity.
%In their analysis, the weight for LS loss $w_{ij}^+ = \frac{1}{ \sum_{k \in \P_i^+} e^{ S_{ij} - S_{ik} } }$ and $w_{ij^-} =  \frac{1}{ \sum_{k \in \N_i^+} e^{ S_{ij} - S_{ik} } }$.
%% \textcolor{blue}{
%% Pair weighting representation of LS loss and MS loss are provided in~\cite{wang2019multi}, which reformulate the original LS loss~\ref{equ:LS}, MS loss~\ref{equ:MS} into a combination of $S_{ij}$ with loss specified weight $\w_{ij}$ of each pair. Pairs have different weighting scheme $\w_{ij}^+, \w_{ij}^-$ in different losses.
%% }
%When setting $\gamma_i^+ = \gamma_i^- = 1$, $p_{ij}^{+*}$ and $p_{ij}^{-*}$ are exactly the $\w_{ij}^+$ , $\w_{ij}^-$ of LS loss (The details of $\w_{ij}^+, \w_{ij}^+$ for both LS and MS are provided in the supplementary).
%This implies that our DRO formulation is able to recover the LS loss.
%%When $\P_i^+ \neq \P_i$ and $\N_i^+ \neq \N_i$, DRO framework recovers the LS loss by additionally applying a threshold pair mining method.
%The details of deriving the closed form of $p_{ij}^{+*}, p_{ij}^{-*}$ are provided in the supplementary.
%Since $\gamma_i^+, \gamma_i^-, i\in {1\cdots n}$ are able to set different values to different anchor $x_i$, our formulation is more flexible than LS loss.

Finally, we can recover MS loss by showing the gradient equivalence between a variant of DRO formulation and MS loss  in a very similar way. %and further combining with some modifications on the uncertainty set $\mathcal{U}$. 
The details  are provided in the supplementary.

% \resizebox{\textwidth}{!} 
% {
% $\max_{ \p_i^+, \p_i^-, i \in \{ 1, \cdots, B \} }  - \sum_{i=1}^{n}(\sum\limits_{j\in \P_i^+ \cup \N_i^+} p_{ij} l_{ij}(\theta) ) + \gamma_i^+ D_{KL}((\p_i^+,p_{i,|\P_i^|+1})|| \frac{\textbf{1}}{|\P_i^+|+1}) + \gamma_i^- D_{KL}((\p_i^-, p_{i,|\N_i^+|+1}) || \frac{\textbf{1}}{|\N_i^+|+1}))) \\
% $
%  }

% \subsection{Classes-Based DRO}
% In anchor based sampling method, each class is assumed to have equal contribution to learning the objective. However, we should put more efforts on the classes that have larger losses.
% Assume we have $C$ class in a batch, $P_c, N_c$ denotes all the positive and negative pairs for each class.

% \begin{equation}
% \centering
% \begin{aligned}
% l_c &= \sum_{i\in \{P_c,N_c\}}l_i(\theta)\\
% p(c) & = \frac{l_c}{\sum_{c=1}^C l_c}
% \end{aligned}
% \end{equation}

\section{Experiments}
\vspace*{-0.1in}Our methods was implemented by Pytorch and using BN-Inception network~(\cite{ioffe2015batch}) pre-trained on ImageNet ILSVRC~(\cite{russakovsky2015imagenet}) to  fairly compare with other works. The same as (\cite{wang2019multi}), a FC layer on the top of the model structure following the global pooling layer was added with randomly initialization for our task. Adam Optimizer with $1e^-5$ learning rate was used for all our experiments. 
 
We verify our DRO framework on image retrieval task with three standard datasets, Cub-200-2011, Cars-196 and In-Shop.
These three datasets are split according to the standard protocol. For Cub-
200-2011, the first 100 classes with 5864 images are used for training, and the the other 100 classes
with 5924 images are saved for testing. Cars-196 consists of 196 car models with 16,185 images.
We use the first 98 classes with 8054 images for training and the remaining 98 classes with 8,131
images for testing. For In-Shop, 997 classes with 25882 images are used for training. The test set
is further partitioned to a query set with 14218 images of 3985 classes, and a gallery set containing
3985 classes with 12612 images. %Detailed descriptions are provided in the supplementary.
Batches are constructed with the following rule: we first sample a certain number of classes and then randomly sample $M$ instances for each class. The standard recall$@k$ evaluation metric is used in all our experiments, where $k = \{1, 2, 4, 8, 16, 32\}$ on Cub-200-2011 and Car-196, and $k = \{1, 10, 20, 30, 40, 50\}$ on In-Shop. 
We apply margin loss ($\mathcal L_M$) and binomial loss ($\mathcal L_B$, \cite{yideep}) as base losses for our DRO framework. 
 $m$ is the margin in $\mathcal L_M$. $\lambda$ is the threshold for both $\mathcal L_M$ and $\mathcal L_B$.
$\alpha$ and $\beta$ are hyperparameters in $\mathcal L_B$.

\subsection{Quantitative Results}
\label{sec:Quantitative Results}
\vspace*{-0.1in}
 
% \begin{table}[t]\label{tab:SOPresults}
% \centering 
% \begin{tabular}{c|cccccc}
% \hline
% Recall$@K$ & 1&10&20&30 & 40 & 50\\ \hline
% FashionNet(\cite{liu2016deepfashion}) &53.7 &73.0 & 76.0 & 77.0 & 79.0 & 80.0 \\ 
% HDC(\cite{oh2017deep})  & 62.1 & 84.9 & 89.0 & 91.2 & 92.3 & 93.1 \\
% HDL(\cite{ge2018deep})&80.9 & 94.3 &  95.8 &  97.2 & 97.4 & 97.8 \\ \hline
% ABIER(\cite{opitz2018deep}) & 83.1& 95.1 &96.9 &97.5 &97.8& 98.0 \\ 
% ABE(\cite{yu2018hard}) &
% 87.3 & 96.7 & 97.9 & 98.2 & 98.5 &98.7 \\ \hline

% MS(\cite{wang2019multi})&89.7 & 97.9 & 98.5  & \textbf{98.8}  &\textbf{99.1}  & \textbf{99.2} \\ \hline
% DRO-TopK$_M$(Ours)&\textbf{91.0}  &\textbf{98.1}& \textbf{98.7} & \textbf{99.0} &\textbf{99.1 } & \textbf{99.2}\\ 
% DRO-TopK$_B$(Ours)& \textbf{90.7}  & 97.7 & 98.4 & \textbf{98.8} & 99.0 & 99.1\\ 
% DRO-TopK-PN$_M$(Ours)& \textbf{91.3}  & \textbf{98.0} & \textbf{98.7} & \textbf{98.9} & \textbf{99.1} & \textbf{99.2}\\
% DRO-TopK-PN$_B$(Ours)& \textbf{91.1}  & \textbf{98.1} & \textbf{98.6} & \textbf{98.8}  &99.0 & \textbf{99.2}\\ 
% DRO-KL$_M$(Ours) &\textbf{90.8}  &
% \textbf{98.0}&\textbf{98.6} & \textbf{99.0} &\textbf{99.1}  &\textbf{99.2} \\ \hline
% %DRO+Margin Loss-G&\textbf{-}  &\textbf{-}& \textbf{-} & \textbf{-}& \textbf{-}& \textbf{-}\\ \hline
% \end{tabular}
% \caption{Recall$@k$ on In-Shop}
% \label{tab:SOTA_Shop}
% \end{table}
\setlength{\textfloatsep}{5pt}

In this experiment, we compare our DRO framework with other SOTA baselines on Cub-200-2011, Cars-196 and In-Shop, which includes \cite{wang2019multi,yu2018hard,kim2018attention,opitz2018deep,ge2018deep,harwood2017smart,wu2017sampling,oh2017deep}.
Among them, mining-based methods are Clusetring, HDC, Margin, Smart Mining and HDL.
ABIER and ABE are ensemble methods.
HAP2S\_E and MS are sampling-based methods, which are highly related to our methods.
For our DRO framework, 
we test all three variants which are proposed in section~\ref{sec:KeyTechnique}.
We apply two loss functions, margin loss and binomial loss, respectively.
Since DRO $\p$-sampling works on all pairs in a batch, the binomial variant may not directly apply to p-sampling.
Thus, it makes totally five variants of our DRO framework, denoted by 
DRO-TopK$_M$, DRO-TopK$_B$, 
DRO-TopK-PN$_M$, DRO-TopK-PN$_B$
and DRO-KL$_M$, where the subscript $M$ and $B$ represent the variants of our framework using margin loss and binomial loss, respectively.
We set embedding space dimension $d=1024$.
The batchsize is set $B=80$ on Cub-200-2011 and Cars-196, $B=640$ on In-Shop. 
$\gamma$ is tuned from the range $ = \{0.1: 0.2: 0.9 \}$ on all three datasets and $K$ is tuned from $ \{ 160, 200, 240, 280\}$ on Cub-200-2011 and 
Cars-196, and selected from $\{640, 960, 1280, 1600, 1920\}$ on In-Shop. 

\begin{table}[t]
\centering
\caption{Recall$@k$ on Cub-200-2011 and Cars-196}
\vspace{-0.5\baselineskip}
\resizebox{\textwidth}{!} 
{
\begin{tabular}{c|cccccc|cccccc}
\hline
                     & \multicolumn{6}{c|}{Cub-200-2011}                & \multicolumn{4}{c}{Cars-196}                  \\ \hline
\multicolumn{1}{c|}{Recall$@k(\%)$} &  1    &  2    &  4    &  8    &   16   & 32    &    1      &      2   &    4     & 8 &16&32       \\ \hline

 Clusetring(\cite{oh2017deep})    &  48.2 & 61.4& 71.8&81.9 &  - & -
& 58.1 & 70.6&80.3& 87.8& -&-   \\ 
HDC(\cite{oh2017deep}) &  53.6 & 65.7& 77.0&85.6 &  91.5 & 95.5
& 73.7 & 83.2&89.5& 93.8& 96.7&98.4   \\
 Margin(\cite{wu2017sampling})    &  63.6 & 74.4& 83.1&90.0 &  94.2 & -
& 79.6 & 86.5& 91.9& 95.1&97.3&-   \\ 
                     Smart Mining(\cite{harwood2017smart})&    49.8 &62.3 & 74.1 & 83.3 &- & -
& 64.7 & 76.2& 84.2 &90.2 &-& -
\\
HDL(\cite{ge2018deep})& 57.1 &68.8 & 78.7 & 86.5 & 92.5 & 95.5 &81.4 &88.0 &92.7& 95.7 &97.4 & \textbf{99.0} \\
\hline

 ABIER(\cite{opitz2018deep})    &  57.5 & 68.7& 78.3 & 86.2 &  91.9 & 95.5
& 82.0 & 89.0& 93.2 &96.1&97.8&98.7   \\ 
                     ABE(\cite{kim2018attention})&    60.6 &71.5 & 79.8 & 87.4 &- & -
& 85.2 & 90.5& 94.0 &96.1 &-& -           \\ 
  
\hline

 HAP2S$\_$E(\cite{yu2018hard})    &  56.1 & 68.3& 79.2 & 86.9 &  - & -
& 74.1 & 83.5& 89.9 & 94.1&-&-   \\ 
                     MS(\cite{wang2019multi})&    65.7 &77.0 & \textbf{86.3} & 91.3 &94.8 & 97.0
& 84.1 & 90.4& 94.0 &\textbf{96.5}&\textbf{98.0}& 98.9           \\ 
  
\hline
                       DRO-TopK$_M$(Ours) & \textbf{67.4} &  \textbf{ 77.7}&  85.9 & \textbf{91.6} &   \textbf{95.0} &  \textbf{97.3}& \textbf{86.0}  &\textbf{91.7 }& \textbf{95.0} & \textbf{97.3}& \textbf{98.5}& \textbf{99.2}              \\
                       DRO-TopK$_B$(Ours) &\textbf{68.1} & \textbf{ 78.4} &  86.0&  \textbf{91.4}  &  \textbf{95.1} &  \textbf{97.6}  
& \textbf{85.4}  &\textbf{91.0 }& \textbf{94.2} & \textbf{96.5}& \textbf{98.0}& \textbf{99.0}          \\
                      DRO-TopK-PN$_M$(Ours) & \textbf{67.3} &  \textbf{ 77.6}&  85.7 & 91.2 &   \textbf{95.0} &  \textbf{97.7}
& \textbf{86.1}  &\textbf{91.7 }& \textbf{95.1} & \textbf{97.1}& \textbf{98.4}& \textbf{99.1}         \\
                      DRO-TopK-PN$_B$(Ours)  &\textbf{67.6} & \textbf{ 77.9} &  86.0&  \textbf{91.8}  &  \textbf{95.2} &  \textbf{97.7}
& \textbf{86.2}  &\textbf{91.7 }& \textbf{95.8} & \textbf{97.4}& \textbf{98.6}& \textbf{99.3}                \\  
DRO-KL$_M$(Ours)  & \textbf{67.7} &  \textbf{ 78.0}&  86.1 & \textbf{91.8} &  \textbf{95.6} &  \textbf{97.8}
& \textbf{86.4}  &\textbf{91.9}& \textbf{95.4} & \textbf{97.5}& \textbf{98.7}& \textbf{99.3}             \\ \hline 
\end{tabular}}

\label{tab:SOTA_cub_car}
\end{table}

Table~\ref{tab:SOTA_cub_car} and~\ref{tab:SOTA_Shop} report the experiment results. 
We mark the best performer in bold in the corresponding evaluation measure on each column.
For our framework, particularly, we mark those who outperform all other SOTA methods in bold.
% \textbf{The bold results in our method are the values of recall$@$k that higher or equal to all the previous SOTA methods.}
We can see that our five variants outperform other SOTA methods on recall$@1$ on all three datasets.
Particularly on Cars-196, our five variants outperforms other SOTA methods on all recall$@k$ measures.
On Cub-200-2011, DRO-TopK$_B$ achieves a higher recall$@1$ (improving $2.4$ in recall$@1$) than the best SOTA, MS. 
On Cars-196, DRO-KL$_M$ has the best performance, which improves $2.3$ and $1.2$ in recall$@1$ compared to the best non-ensemble SOTA, MS, and the best ensemble SOTA, ABE. 
On In-Shop, DRO-TopK-PN$_M$ improves $1.6$ in recall$@1$ compared to the best results among SOTA, MS.
The above results verify 1) the effectiveness of our DRO sampling methods and 2) the flexibility of our DRO framework to adopt different losses. %\textcolor{red}{Addition experimental results of the mean and variance of recall$@k$ under multiple runs on Cub-200-2011 are provided in the supplementary (Figure~\ref{fig:mean_and_std})}.

\subsection{Ablation Study}
\vspace*{-0.1in}
\subsubsection{Comparison with LS loss and MS loss }

% \begin{table}[htbp]
\begin{table}[htbp]
\centering
\caption{Recover of MS loss and LS loss on Cub-200-2011 and Cars-196}
\vspace{-0.5\baselineskip}
\resizebox{\textwidth}{!} 
{
\begin{tabular}{c|cccccc|cccccc}
\hline
                     & \multicolumn{6}{c|}{Cub-200-2011}                & \multicolumn{4}{c}{Cars-196}                  \\ \hline
\multicolumn{1}{c|}{Recall$@K(\%)$  } &  1    &  2    &  4    &  8    &   16   & 32    &    1      &      2   &    4     & 8 &16&32       \\ \hline
                     MS&   55.6 & 67.7& 77.4 &86.3 & 92.1& 95.8
& 73.2 & 81.5& 87.6 & 92.6   &- & -            \\ 
{\ul }              LS   &  56.8 & 67.9& 77.5 & 85.6 &  91.2 & 95.2
& 69.7 & 79.3& 86.2 & 91.1&-&-                 \\ 
\hline
                       DRO-KL-G-$\gamma=1$ & 56.4 & \textbf{68.3}& \textbf{78.9} & \textbf{86.3} & 
\textbf{91.7}&\textbf{95.8}& 70.5 & 79.8& 86.6 & 91.6 &\textbf{94.9} &\textbf{97.1}                \\
                       DRO-KL-G-$\gamma=0.1$ & 56.8 &  \textbf{68.7}& \textbf{79.0} & \textbf{86.6} &  \textbf{92.1} &\textbf{95.9}
& 72.5 &  \textbf{81.9}& \textbf{88.1} & 92.3 &\textbf{95.4} &\textbf{97.3 }               \\
                      DRO-KL-G-$\gamma=0.01$ & \textbf{57.0} & \textbf{ 69.4}& \textbf{79.9} & \textbf{87.0} & \textbf{92.3} & \textbf{95.9}
& \textbf{73.1} &\textbf{82.2} & \textbf{ 88.8} & \textbf{93.4}      &\textbf{96.2} & \textbf{98.0}         \\
                      DRO-KL-G-$\gamma=0.001$ &  56.7 & \textbf{ 68.5}& \textbf{79.0} & \textbf{87.3} & \textbf{92.6} & \textbf{96.0} 
& \textbf{75.0} &\textbf{83.4} & \textbf{ 89.5} & \textbf{93.7}   & \textbf{96.6} & \textbf{98.3}                \\ \hline 
\end{tabular}}
\label{tab:recover_cub_car}
\vspace*{-0.1in}\end{table}
In Section~\ref{section:recover_LS_loss}, we theoretically show that LS loss and MS loss can be viewed as special cases of our DRO framework.
% is a special case of our DRO framework using margin loss with $\gamma = \gamma_i^+ = \gamma_i^-=1, i \in [B]$, and MS loss is a special case of our DRO framework using margin loss with  $\gamma_i^+ = \frac{1}{\alpha},  \gamma_i^-=\frac{1}{\beta}, i\in [B]$.
In this experiment, we aim to empirically demonstrate that our framework is general enough and recovers LS loss.
Specifically, we would show
1) when $\gamma = 1$, our framework performs similarly to LS loss, as stated in Section~\ref{section:recover_LS_loss},
2) our framework can be seen as a {\it generalized LS loss} by treating $\gamma$ as a hyper-parameter, and
3) our {\it generalized LS loss} outperforms MS loss, even though the performance of the ordinary LS loss is inferior to that of MS loss.

We adopt the set up of embedding dimension and batchsize in the ablation study of~\cite{wang2019multi}, i.e., $d = 64$ and $B=80$. Therefore, we report the existing results of MS and LS loss presented in~\cite{wang2019multi} on Cars-196.
For Cub-200-2011 and In-Shop, we implement MS and LS loss according to \cite{wang2019multi}.
Following \cite{wang2019multi}, we set $\alpha = 2, \beta = 50$ for MS loss.
For our DRO framework, we apply grouping to $\p$ by equation~(\ref{equ:DRO_Lifted_Structure_Loss}), and denote this variant of DRO framework as DRO-KL-G. We set $\gamma_i^+ = \gamma_i^- = \gamma =\{1, 0.1, 0.01, 0.001\}, i\in [B]$ for DRO-KL-G , $m = 0.2$ for the margin loss, and $\lambda = 0.5$ for all three losses (MS, LS and margin loss).
%We conduct experiments on our backbone model with embedding dimension 64 and batch size 80 compare with MS loss and lifted structure loss using the same setting. %Experiment results shows that, by setting different $\lambda$, our DRO framework performance slightly better than MS loss and lifted structure loss. 
% \textcolor{red}{
As the pairs with zero loss will not contribute to the updates of model but affect the calculation of $\p$ in DRO framework, we remove the pairs with zero loss to further promotes training. 
% For fair comparison, we adopt the threshold mining, $i.e \ P = \{ S_{ij}| S_{ij} < \lambda + m, y_{ij} = 1\},N = \{ S_{ij}| S_{ij} > \lambda - m, y_{ij} = -1\}$ to MS loss and LS loss as well.
% }

Table \ref{tab:recover_cub_car} and \ref{tab:recover_inshop} show experiment results on Cub-200-2011, Cars-196 and In-Shop, respectively. 
As can be seen, the performance of MS loss is better than LS loss on three datasets, particularly on Cars-196, which also verifies the results of ablation study in \cite{wang2019multi}.
When $\gamma = 1$, our method performs similarly to LS loss, which verifies that our method recovers LS loss.
Furthermore, when we treat $\gamma$ as a hyper-parameter (especially $\gamma = 0.001$) and regard our framework as {\it generalized LS loss}, our method obtain improved performance compared to the ordinary LS loss.
Lastly, even if MS loss exploits pseudo positive and negative pairs, our {\it generalized LS loss} outperforms MS loss.

\begin{minipage}{\textwidth}
% \resizebox{\textwidth}{!} {
\small{
  \begin{minipage}{0.49\textwidth}
    \centering
    \includegraphics[width =\textwidth]{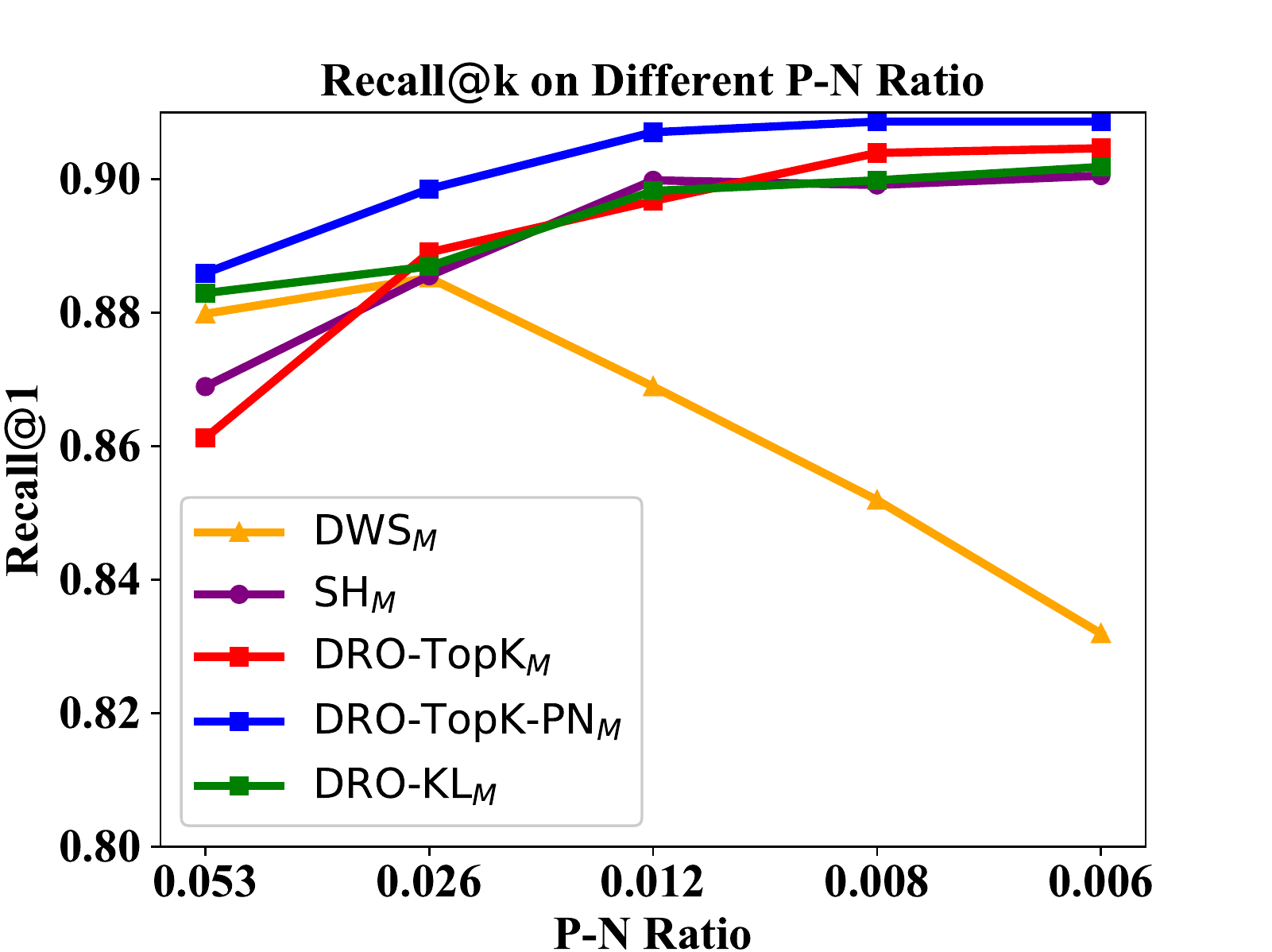}
     \vspace{-0.8\baselineskip}
     \captionof{figure}{Recall vs Imbalance Ratio }
    \label{fig:resisting_imbalance}
  \end{minipage}
  \hfill
  \begin{minipage}{0.49\textwidth}
    \centering
    \captionsetup{font=footnotesize}\captionof{table}{Recall$@k$ on In-Shop}
 \vspace{-1\baselineskip}
    \resizebox{0.9\textwidth}{!}{
\begin{tabular}{c|cccccc}
\hline
Recall$@K$ & 1&10&20&30 & 40 & 50\\ \hline
FashionNet(\cite{liu2016deepfashion}) &53.7 &73.0 & 76.0 & 77.0 & 79.0 & 80.0 \\ 
HDC(\cite{oh2017deep})  & 62.1 & 84.9 & 89.0 & 91.2 & 92.3 & 93.1 \\
HDL(\cite{ge2018deep})&80.9 & 94.3 &  95.8 &  97.2 & 97.4 & 97.8 \\ \hline
ABIER(\cite{opitz2018deep}) & 83.1& 95.1 &96.9 &97.5 &97.8& 98.0 \\ 
ABE(\cite{yu2018hard}) &
87.3 & 96.7 & 97.9 & 98.2 & 98.5 &98.7 \\ \hline

MS(\cite{wang2019multi})&89.7 & 97.9 & 98.5  & \textbf{98.8}  &\textbf{99.1}  & \textbf{99.2} \\ \hline
DRO-TopK$_M$(Ours)&\textbf{91.0}  &\textbf{98.1}& \textbf{98.7} & \textbf{99.0} &\textbf{99.1 } & \textbf{99.2}\\ 
DRO-TopK$_B$(Ours)& \textbf{90.7}  & 97.7 & 98.4 & \textbf{98.8} & 99.0 & 99.1\\ 
DRO-TopK-PN$_M$(Ours)& \textbf{91.3}  & \textbf{98.0} & \textbf{98.7} & \textbf{98.9} & \textbf{99.1} & \textbf{99.2}\\
DRO-TopK-PN$_B$(Ours)& \textbf{91.1}  & \textbf{98.1} & \textbf{98.6} & \textbf{98.8}  &99.0 & \textbf{99.2}\\ 
DRO-KL$_M$(Ours) &\textbf{90.8}  &
\textbf{98.0}&\textbf{98.6} & \textbf{99.0} &\textbf{99.1}  &\textbf{99.2} \\ \hline
%DRO+Margin Loss-G&\textbf{-}  &\textbf{-}& \textbf{-} & \textbf{-}& \textbf{-}& \textbf{-}\\ \hline
\end{tabular}}
\vspace*{0.05in}
\label{tab:SOTA_Shop}
   \captionsetup{font=footnotesize}
   {\captionof{table}{Recover of MS loss and LS loss on In-Shop}}
   \vspace{-1\baselineskip}
         \resizebox{0.9\textwidth}{!}{
\begin{tabular}{c|cccccc}
\hline
Recall$@K(\%)$& 1&10&20&30 & 40 & 50\\ \hline
MS & 79.8 & 94.9& 96.8 & 97.6 & 97.9& 98.3\\ 
LS& 82.6 & 94.1& 95.6 & 96.4 &96.9& 97.4\\ \hline
DRO-KL-G-$\gamma = 1$ & \textbf{84.8} &  \textbf{95.9}& \textbf{97.3} & \textbf{97.9} & \textbf{98.2} & \textbf{98.5}\\ 
DRO-KL-G-$\gamma = 0.1$ & \textbf{85.1} &  \textbf{96.1}& \textbf{97.5} & \textbf{98.0} &  \textbf{98.3}& \textbf{98.5}\\ 
DRO-KL-G-$\gamma = 0.01$ & \textbf{85.8} & \textbf{96.2} &  \textbf{97.9}& \textbf{97.8} & \textbf{98.2} & \textbf{98.4}\\ 
DRO-KL-G-$\gamma = 0.001$ & \textbf{85.7} & \textbf{96.1} & \textbf{97.4} & \textbf{97.9} & \textbf{98.2} & \textbf{98.5}\\\hline
\end{tabular}}
   
\label{tab:recover_inshop}
    \end{minipage}}
  \end{minipage}

\vspace*{-0.1in}
\subsubsection{Capacity to Handle Pair Imbalance.}
\label{sec:Capacity to Handle Pair Imbalance.}
\vspace*{-0.1in}\label{section:experiment_imbalance}
In this experiment, we compare our DRO framework with different sampling methods, i.e., semihard (SH) and DWS, in terms of sensitivity to the imbalance ratio. By setting different batchsizes $B\in \{80, 160, 320, 480, 640\}$, we have different positive-negative pair ratios $|\P|:|\N| \in \{ 0.053, 0.026, 0.012, 0.008, 0.006 \}$.
For all methods, we apply margin loss and set $M = 5$ for each class and embedding space dimension $d = 1024$.
SH mining is originally designed for triplet loss.
Since there is no straightforward choice for the positive pair, we use $\lambda$ as the upper bound to simulate the similarity of the positive pair in triplet loss. 
For DWS, we follow the parameter setting in the original paper~(\cite{wu2017sampling}).
We apply margin loss in the proposed three variants our DRO framework, which are denoted by DRO-TopK$_M$, DRO-TopK-PN$_M$ and DRO-KL$_M$, respectively.
We set $K = 2*B$ both for DRO-TopK$_M$, DRO-TopK-PN$_M$. 
We evaluate recall$@1$ of all methods and report results in Figure \ref{fig:resisting_imbalance}.

Figure~\ref{fig:resisting_imbalance} shows that the DWS has better performance when the positive-negative pair ratio is relatively large, and encounters a sharp decrease in recall$@1$ when the ratio decreases.
Other four methods obtain better performance when the positive-negative ratio increases.
Among them, DRO-TopK$_M$ and DRO-KL$_M$ have similar performance, with SH on all positive-negative pair ratios,
while they perform slightly better than SH when the positive-negative ratios are small.
DRO-TopK-PN$_M$ constantly outperforms all other methods.
The reason why DWS performs poorly when the positive-negative pair ratio is small may be that DWS aims to sample pairs uniformly in terms of distance (\cite{wu2017sampling}), while our DRO framework and SH focus more on hard pairs.
To sum up, our framework achieves very competitive performance against SOTA methods,
and maintains increasing recall$@1$ as the positive-negative ratio increases.
These two observations together demonstrate the effectiveness of our DRO framework to handle pair imbalance. 
%\textcolor{red}{Trends of recall$@k$ on embedding space dimension 512 is provided in the supplementary (Figure~\ref{fig:runtime-imbanlance-512}) to show that the ability of our methods to handle imbalance pairs is not sensitive to network architecture.}

\vspace*{-0.06in}
\subsubsection{Sensitivity of $K$ in Top-K}
\vspace*{-0.1in}As we mentioned in section~\ref{sec:introduction}, selecting too many pairs within a batch will leads to poor performance of the model.
On the other hand, when the number of selected pairs is too small, the model would be sensitive to outliers. 
In this experiment, we study the sensitivity of $K$ in our DRO framework--how the performance of our DRO framework is affected by the value of $K$.
%Furthermore, SH and DWS have relatively good performance by using the number of pairs which twice number of positive pairs (\cite{schroff2015facenet}, \cite{wu2017sampling})..
%the number of negative pairs equal to the positive pairs in previous studies 
We set the batchsize $B=640$ and $M = 5$, which makes the number of positive pairs $|\P|=1280$ and the number of negative pairs $|\N| = 198080$.
We set $K$ from the range $\{640, 960, 1280, 1600, 1920, 2560 \}$ and evaluate recall$@k$ for models trained by different $K$.
We choose the above range of $K$ according to the number of pairs selected by DWS and SH in Section \ref{section:experiment_imbalance} (both selects $2560$ pairs roughly).

\begin{figure}[t]
   
    \begin{minipage}[t]{0.49\textwidth}
     \centering
    \includegraphics[width=0.8\textwidth]{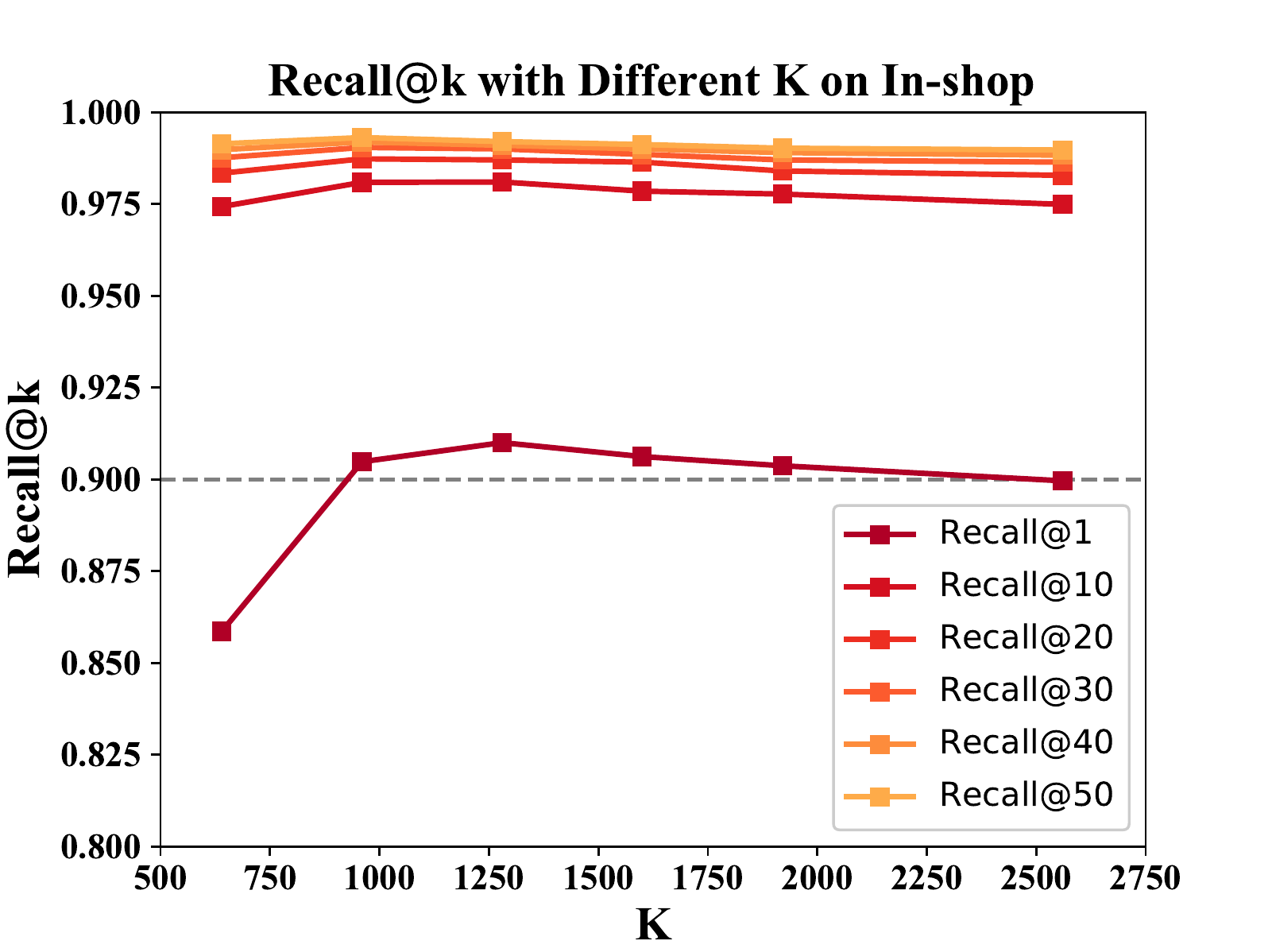}
    
   \vspace{-0.8\baselineskip}
    \caption{The effects of K on recall$@k$ on In-Shop}
    \label{fig:sensitivityofK}
    \end{minipage}
    \begin{minipage}[t]{0.49\textwidth}
     \centering
    \includegraphics[width=0.8\textwidth]{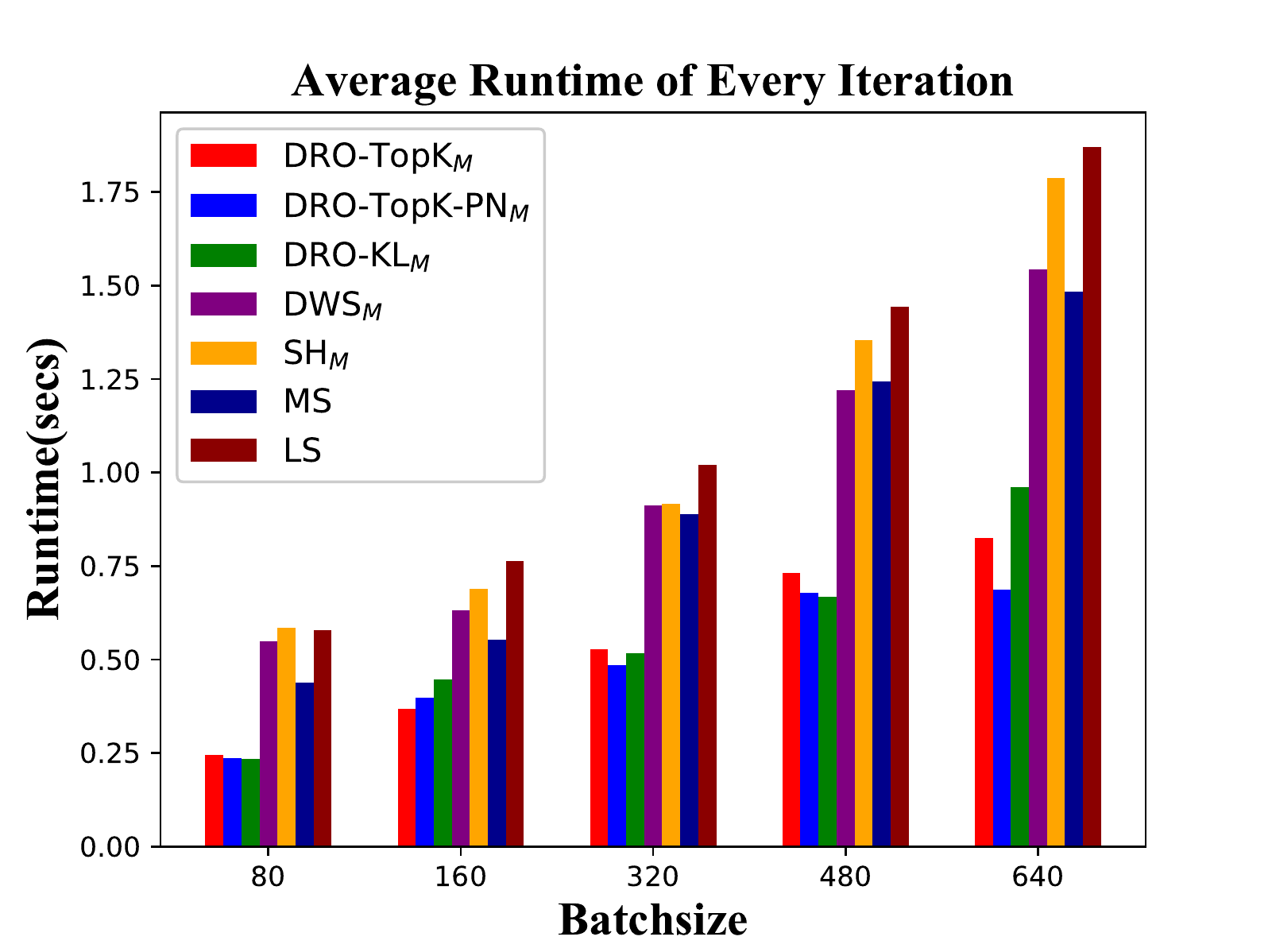}
    
    \vspace{-0.8\baselineskip}
    \caption{Average running time of every iteration}
    \label{fig:run-time}
    \end{minipage}
\end{figure}

Figure~\ref{fig:sensitivityofK} illustrates how different values of $K$ affect recall$@k$ on In-Shop. 
We can see that, DRO-TopK$_M$ performs best when $K = 1280$ and recall$@k$ is stable on the entire range of $K$.
Our DRO framework is not sensitive to $K$ when $K$ is in a reasonably large range.

 \subsubsection{Runtime Comparison}
\vspace*{-0.1in}
Next, we compare the running time of our proposed three variants of our DRO framework with different pair mining methods, MS and LS losses on In-shop. 
Our experiments conducted on eight GTX1080Ti GPU.
The embedding dimension $d = 1024$, and results are compared under different batchsize $B = \{80, 160, 320, 480, 640\}$. The same as previous experiments, we set $K = 2*B$ both for DRO-TopK$_M$ and DRO-TopK-PN$_M$. $\gamma = 0.1$ for DRO-KL$_M$.
SH is implemented according to the  paper~\cite{schroff2015facenet},~\cite{wu2017sampling}.
DWS and MS are implemented based on the code provided by the author. 
LS loss is implemented following the code provided by \cite{wang2019multi}.

Figure~\ref{fig:run-time} reports the average running time of each iteration on 200 epochs. 
We can see that all of three proposed variants of DRO framework run faster than other {\it{anchor-based}} mining methods and losses. 
For all of our three variants, pairs are selected directly from all the pairs,
while additional cost is required to select pairs anchor by anchor in other methods. 
LS loss is slower than MS loss, because MS mining is applied to MS loss,
which would reduce the number of pairs for computing subgradients when updating the model.
For DWS, the distance distribution of negative pairs is only calculated once for each iteration. 
It thus only needs to select pairs according to the pre-computed distribution for each anchor.
In contrast, SH requires to compare negative pairs with the lower and upper bound of an interval at each iteration for each anchor, which increases the computational burden.
% \textcolor{red}{additional computation is required for the negative pairs of each anchor to compare with the margin and the positive pair similarity simulating threshold $\lambda$}.
It can be the reason why SH is slower than DWS.

\section{Conclusion}
\vspace*{-0.1in}In this paper, we cast DML as a simple pairwise binary classification problem and formulate it as a DRO framework.
Compared to existing DML methods that leverage all pairs in a batch or employ heuristic approaches to sample pairs, our DRO framework constructs a robust loss to sample informative pairs, which also comes with theoretical justification from the perspective of learning theory.
Our framework is general since it can include many novel designs in its uncertainty decision set.
Its flexibility allows us to recover the state-of-the-art  loss functions.
Experiments show that our framework outperforms the state-of-the-art DML methods on benchmark datasets.
We also empirically demonstrate that our framework is efficient, general and flexible.

% provide a unified simple and effective DRO framework for DML. 
% First, our DRO framework is general enough to recover the SOTA pair-based loss functions and flexible enough to collaborate with different base loss functions.
% In addition, compared to existing methods, \textcolor{blue}{we verify that our DRO framework particularly addresses the pair imbalance issue in DML}.
% Last but not least, DRO framework outperforms SOTA DML methods in terms of recall$@k$ on three benchmark image retrieval dataset, which dmonstrates the efficacy of our framework.

\bibliography{iclr2019_conference}
\bibliographystyle{iclr2019_conference}

% Authors must not appear in the submitted version. They should be hidden
% as long as the \iclrfinalcopy macro remains commented out below.
% Non-anonymous submissions will be rejected without review.

\section{Supplementary}

\subsection{Derivation of Recover SOTA LOSSES}

In this section, we show how our DRO framework recovers SOTA loss functions, LS loss and MS loss.
Their definitions are as follows, respectively.
\begin{align}
\label{equ:LS-ori}
\mathcal L_{LS} = \sum_{i=1}^{B} [\log\sum_{k\in P_i}e^{\lambda - S_{ik}} + \log\sum_{k\in N_i}e^{S_{ik}-\lambda}]_+  ,
\end{align}
where $\lambda$ is the margin hyper-parameter.
\begin{align}
\label{equ:Ms-ori}
\mathcal L_{MS} = \frac{1}{n}\sum_{i=1}^{n}\{\frac{1}{\alpha}\log[1 + \sum_{k\in P_i}e^{-\alpha(S_{ik}-\lambda)}] + \frac{1}{\beta}\log[1 + \sum_{k\in N_i}e^{\beta(S_{ik}-\lambda)}]\}
\end{align}
where $\lambda$ is the margin hyper-parameter and $\alpha$ and $\beta$ are coefficient hyper-parameters.
%\textcolor{red}{where equality of (a) is due to
%$p_{ij}^{+*} = w_{ij}^+$ and $p_{ij}^{-*} = w_{ij}^-$.
%$C$ is a constant that absorbs all constants, so it has no influence on the updates of $\theta$. 
%}

\subsubsection{LS Loss under Our Framework}

Recall that the objective function is decomposable in terms of $\p_i = [\p_i^+, \p_i^-] \in \mathbb R^{P_i + N_i}$. 
We denote the $p_{ij}= p_{ij}^+$ when $j\in\P_i$, and $p_{ij}= p_{ij}^-$ when $j \in \N_i$ for simplicity.
The Lagrangian function of~(\ref{equ:DRO_Lifted_Structure_Loss}) can be represented as:
\begin{equation}
    \begin{aligned}
    \mathcal{L}(\p,  \mathbf{v}) & = \sum\limits_{i=1}^B \mathcal{L}(\p_i, v_i^+, v_i^-)  ,
    \end{aligned}
\end{equation}
where
\begin{equation}
\label{equ:Lagrange}
    \begin{aligned}
    \mathcal{L}(\p_i, v_i^+, v_i^-) &=   -\sum_{j\in \P_i \cup \N_i} p_{ij} l_{ij}(\theta) + \gamma_i^+ D_{KL}(\P_i || \frac{\textbf{1}}{|\P_i|}) + \gamma_i^- D_{KL}(\p_i^- || \frac{\textbf{1}}{|\N_i|}) \\
    & + v_i^+(\sum_{j \in \P_i} p_{ij} -1) + v_i^-(\sum_{j \in \N_i} p_{ij} -1)  .
   \end{aligned}
\end{equation}
% where $\p_i = [\P_i, \p_i^-] \in R^{|\P_i\cup\N_i|}$
% \begin{equation}
% \label{equ:Lagrange}
% \centering
%     \begin{aligned}
%     \max\limits_{v_i^+,v_i^-} \min\limits_{\p_i}\mathcal{L}(\p_i, v_i^+, v_i^-) 
%      \end{aligned}
% \end{equation}
According to KKT conditions, $v_i^{+*}$ and  $v_i^{-*}$ are the optimal solutions of the dual function, 
and $\p_i^*$ is the optimal solution of the primal problem~(\ref{equ:DRO_Lifted_Structure_Loss}), if and only if
%\begin{equation}
    \begin{align}
    &\frac{d\mathcal{L}}{d\p_i^*} = 0   ,   \label{equ:first-optimality-condition}\\
    & v_i^{+*}(\sum_{j \in \P_i} p_{ij}^{+*} -1) + v_i^{-*}(\sum_{j \in \N_i}      p_{ij}^{-*} -1) = 0     .\label{equ:slack-condition}
   \end{align}
%#\end{equation}

We first derive $\p_i^*$ in terms of $v_i^+, v_i^-$ using equation~(\ref{equ:first-optimality-condition}), i.e.: 
\begin{equation}
\label{equ:KKTopt1}
    \begin{aligned}
\frac{d\mathcal{L}}{d\p_i^*} = -\mathbf{l}_i(\theta)+ \gamma_i^+ \log(n\p_i^{+*})+\gamma_i^- \log(n\p_i^{-*})+\gamma_i^+ \1^+ +\gamma_i^- \1^-+ v_i^+ \1^+ + v_i^- \1^- = 0 \\
\end{aligned}
\end{equation}
where $\mathbf{l}_i(\theta) = \{l_{i1}(\theta),\cdots, l_{i|p_i^+\cup N_i^+|}(\theta) \},\1^+\in R^{|\P_i|}, \1^-\in R^{|\N_i|}$. 
Then the closed form of $\p_i^*$ for positive pairs and negative pairs can be written as follows
\begin{equation}
\label{equ:KKT1solution}
\begin{aligned}
p_{ij}^{+*} = \frac{1}{|\P_i|}e^{\frac{l_{ij}(\theta) - v_i^{+*}}{\gamma_i^+}-1}
,
p_{ij}^{-*} = \frac{1}{|\N_i|}e^{\frac{l_{ij}(\theta) - v_i^{-*}}{\gamma_i^-}-1}  .
\end{aligned}
\end{equation}

Substitute $p_{ij}^{+*}, p_{ij}^{-*}$ into equation~(\ref{equ:slack-condition}), which means $v_i^{+*}$ and $v_i^{-*}$ need to satisfy:
\begin{equation}
\label{equ:v_slack_condition}
\begin{aligned}
 v_i^{+*}(\sum_{j \in \P_i}\frac{1}{|\P_i|}e^{\frac{l_{ij}(\theta) - v_i^{+*}}{\gamma_i^+}-1} -1) + v_i^{-*}(\sum_{j \in \N_i} \frac{1}{|\N_i|}e^{\frac{l_{ij}(\theta) - v_i^{-*}}{\gamma_i^-}-1} -1) = 0  .
\end{aligned}
\end{equation}
Even though equal~(\ref{equ:v_slack_condition}) also equals to 0 when $v_i^{+*} = 0 $ or $v_i^{-*} = 0$, or $v_i^{+*} = v_i^{-*} = 0$, but the corresponding optimal solution $\p_i^*$ will not
meet the equality constraints, i.e., $\sum_{j \in \P_i} p_{ij}^{+*} = 1$ and
$\sum_{j \in \P_i} p_{ij}^{+*} = 1$, in the original formulation~(\ref{equ:DRO_Lifted_Structure_Loss}).
Therefore, we only have
\begin{equation}
\label{equ:close_form_of_v_i}
\begin{aligned}
 \sum_{j \in \P_i}\frac{1}{|\P_i|}e^{\frac{l_{ij}(\theta) - v_i^{+*}}{\gamma_i^+}-1} = 1,~~~
 \sum_{j \in \N_i} \frac{1}{|\N_i|}e^{\frac{l_{ij}(\theta) - v_i^{-*}}{\gamma_i^-}-1} = 1  .
\end{aligned}
\end{equation}

Then from equation~(\ref{equ:close_form_of_v_i}), we can get
\begin{equation}
    \begin{aligned}
    v_i^{+*} = \gamma_i^+ \log \Big( \sum_{j \in \P_i}\frac{1}{|\P_i|}e^{\frac{l_{ij}(\theta)}{\gamma}-1} \Big) , ~~~
    v_i^{-*} =  \gamma_i^-  \log \Big( \sum_{j \in \N_i} \frac{1}{|\N_i^-|}e^{\frac{l_{ij}(\theta)}{\gamma}-1} \Big)   .
    \end{aligned}
\end{equation}
Plugging them into~(\ref{equ:KKT1solution}) and apply margin loss as the base loss function for each pair, $l_{ij}(\theta) = [m + y_{ij}( \lambda - S_{ij} )]_+$, we successfully derive the weighting representation of LS loss:
\begin{equation}
\label{equ:close-form-LS}
\centering
\begin{aligned}
p_{ij}^{+*} &= \frac{e^{\frac{l_{ij}(\theta)}{\gamma}}}{\sum\limits_{k\in \P_i}e^{\frac{l_{ik}(\theta)}{\gamma}}}
\stackrel{y_{ij}=1}{=}\frac{e^{\frac{[m +  ( 
\lambda-S_{ij})]_{+}}{\gamma}}}{\sum\limits_{k\in \P_i}e^{\frac{[\alpha + [\lambda -S_{ik})]_{+}}{\gamma}}}
= \frac{1}{\sum\limits_{k\in \P_i}e^{\frac{S_{ij}-S_{ik}}{\gamma}}} \\
p_{ij}^{-*} &= \frac{e^{\frac{l_{ij}(\theta)}{\gamma}}}{\sum\limits_{k\in \N_i}e^{\frac{l_{ik}(\theta)}{\gamma}}}
\stackrel{y_{ij}=-1}{=}\frac{e^(\frac{[m + (S_{ij} - \beta)]_{+}}{\gamma})}{\sum\limits_{k\in \N_i}e^{\frac{[\alpha + (S_{ik} - \beta)]_{+}}{\gamma}}}
= \frac{1}{\sum\limits_{k\in \N_i}e^{\frac{S_{ik}-S_{ij}}{\gamma}}} \\
\end{aligned}
\end{equation}
%\textcolor{red}{We can see that  $p_{ij}^{+*}$ equals to $\w_{ij}^+$ in equation~(\ref{equ:w-ij+LS}),
%$p_{ij}^{-*}$ equals to $\w_{ij}^-$ in equation~(\ref{equ:w-ij-LS}) when $\gamma_i^+ = \gamma_i^- = 1$.}
Thus, when updating the model parameter $\theta$, we are going to minimize the following objective function:
\begin{equation}
\label{equ:weight-form}
    \begin{aligned}
g(\theta,\p^*) &= \sum_{i=1}^{B} \sum_{j \in \P_i \cup \N_i} p_{ij}^{*} l_{ij}(\theta)\\
&= \sum_{i=1}^{B} \Big(\sum_{j \in \P_i} p_{ij}^{+*} l_{ij}(\theta) +\sum_{j \in \N_i} p_{ij}^{-*} l_{ij}(\theta) \Big) \\
%&\mathop{=}\limits_{}^{(a)}\sum_{i=1}^{B} \Big(\sum_{j =1}^B \w_{ij}^- S_{ij} - \sum_{j =1}^B \w_{ij}^+ S_{ij} \Big) + C
\end{aligned}
\end{equation}
Taking the gradients to equation~\ref{equ:weight-form} in terms of $\theta$, we can get:
\begin{equation}
\label{equ:gradients-of-g-2}
\begin{aligned}
\frac{\partial g(\theta,\p^*)}{\partial \theta} &= \sum_{i=1}^{B} \Big( \sum_{j \in \P_i} -p_{ij}^{+*} \frac{\partial S_{ij}}{\partial \theta} + \sum_{j \in \N_i}^B p_{ij}^{-*} \frac{\partial S_{ij}}{\partial \theta} \Big)
\end{aligned}
\end{equation}
Similarly, we take gradients to the $\mathcal L_{LS}$ loss function~(\ref{equ:LS-ori}):

\begin{equation}
\label{equ:gradients-of-LS}
\begin{aligned}
\frac{\partial \mathcal L_{LS}}{\partial \theta} &= \sum_{i=1}^{B} \Big(\sum_{j \in \P_i} \frac{-1}{\sum\limits_{k\in \P_i}^B e^{S_{ij}-S_{ik}}} \frac{\partial S_{ij}}{\partial \theta} + \sum_{j \in \N_i}^B \frac{1}{\sum\limits_{k\in\N_i}^B e^{S_{ik} -S_{ij}}} \frac{\partial S_{ij}}{\partial \theta} \Big)
\end{aligned}
\end{equation}

By substituting equation~(\ref{equ:close-form-LS}) into equation~(\ref{equ:gradients-of-g}), we can see that equation~(\ref{equ:gradients-of-LS}) and~(\ref{equ:gradients-of-g}) are equivalent. This shows our DRO framework successfully recovers the LS loss by setting the uncertainty decision set $\mathcal{U}$ in equation~(\ref{equ:DRO_Lifted_Structure_Loss}).

%\textcolor{red}{where equality of (a) is due to
%$p_{ij}^{+*} = w_{ij}^+$ and $p_{ij}^{-*} = w_{ij}^-$.
%$C$ is a constant that absorbs all constants, so it has no influence on the updates of $\theta$. 
%}

\subsubsection{MS Loss under Our Framework}
MS loss, a combination of binomial loss and LS loss, can also be formulated into our DRO framework.
LS loss only considers the lifted structure between pairs, while binomial loss focusing on the intrinsic property of an independent pair while encoding the pair class information.
% \textcolor{red}{
% From the DRO perspective, the intrinsic property of a pair in binomial loss can be preserved by introducing an additional by adding $\x_i$ into $\P_i$ and $\N_i$, and denote as $\P_i + 1$ and $\N_i +1 $ respectively. 
% }
To recover MS loss under our framework, we re-define $\p \in [0, 1]^{P^+ + N^+ + 2B}$ by adding one more element to $\P_i$ and $\p_i^-$.
Therefore, now we have $\p_i^+ \in [0, 1]^{P_i^+ + 1}$ and $\p_i^- \in [0, 1]^{N_i^+ + 1}$, where the newly added element corresponds to a zero loss, and thus does not contribute to the computation of overall loss.
Then based on the formulation of LS loss, the formulation of MS loss can be written as:
\begin{equation}
\label{equ:DRO_MS_Loss}
\centering
\begin{aligned}
% \max_{\P_i, \p_i^-, i \in [B]}
\max_{\p \in [0, 1]^{P^+ + N^+ + 2B} }
~&~
\sum_{i=1}^B \Big( \sum_{j \in \mathcal P_i \cup \mathcal N_i} p_{ij} l_{ij} (\theta) 
- \gamma_i^+ D_{KL} ( \P_i \| \frac{1}{ |\mathcal P_i|+1 } ) 
- \gamma_i^- D_{KL} (\p_i^- \| \frac{1}{ |\mathcal N_i|+1 } )  \Big)
\\
\text{s.t.}
~&~
\sum_{ j \in \P_i + 1 } p_{ij} = 1, 
\sum_{ k \in \N_i + 1 } p_{ik} = 1, i \in [B]   .
\end{aligned}
\end{equation}
Note that 
\begin{equation}
     \begin{aligned}
      \sum_{j \in \mathcal P_i \cup \mathcal N_i} p_{ij} l_{ij} (\theta) 
      + p_{i, P_i^+}^+ \cdot 0 
      + p_{i, N_i^+}^- \cdot 0
      = \sum_{j \in \mathcal P_i + 1\cup \mathcal N_i +1} p_{ij} l_{ij} (\theta), \forall i\in [B]   .
     \end{aligned}
\end{equation}
% Since $l_{ii}(\theta) = 0,\forall i\in [B]$.

As the analysis of LS loss, we can also obtain the representation of MS loss under our DRO framework from formulation~(\ref{equ:DRO_MS_Loss}), i.e.,
\begin{equation}
\label{equ:closed-form-MS}
\begin{aligned}
p_{ij}^{+*} &= \frac{e^{\frac{l_{ij}(\theta)}{\gamma_i^+}}}{e^{\frac{l_{i,|\P_i|+1}}{\gamma_i^+}}+ \sum\limits_{k\in P_i}e^{\frac{l_{ik}}{\gamma_i^+}}}
\stackrel{y_{ij}=1}{=}\frac{e^{\frac{[m + (\lambda - S_{ij})]_{+}}{\gamma_i^+}}}{1+\sum\limits_{k\in \P_i}e^{\frac{[m + (\lambda - S_{ik})]_{+}}{\gamma_i^+}}}
= \frac{1}{e^{\frac{S_{ij}-c^+}{\gamma_i^+}} + \sum\limits_{k\in \P_i}e^{\frac{S_{ij} -S_{ik}}{\gamma_i^+}}}    ,
\\
p_{ij}^{-*} &= \frac{e^{\frac{l_{ij}}{\gamma_i^-}}}{e^{\frac{l_{i,|\N_i|+1}}{\gamma_i^-}}+\sum\limits_{k\in \N_i}e^{\frac{l_{ik}}{\gamma_i^-}}}
\stackrel{y_{ij}=-1}{=}\frac{e^{\frac{[m +  (S_{ij} - \lambda)]_{+}}{\gamma_i^-}}}{1+\sum\limits_{k\in \P_i}e^{\frac{[m +  (S_{ik} - \lambda)]_{+}}{\gamma_i^-}}}
= \frac{1}{e^{\frac{c^- - S_{ij}}{\gamma_i^-}}+\sum\limits_{k\in \N_i}e^{\frac{S_{ik}-S_{ij}}{\gamma_i^-}}}    ,
\end{aligned}
\end{equation}
where $c^+, c^-, \gamma_i^+, \gamma_i^-$ are hyperpatmeters.
Similar to LS loss, waking the gradients to $\partial g(\theta,\p^*)$ in terms of $\theta$, we can get:
\begin{equation}
\label{equ:gradients-of-g}
\begin{aligned}
\frac{\partial g(\theta,\p^*)}{\partial \theta} &= \sum_{i=1}^{B} \Big( \sum_{j \in \P_i} -p_{ij}^{+*} \frac{\partial S_{ij}}{\partial \theta} + \sum_{j \in \N_i}^B p_{ij}^{-*} \frac{\partial S_{ij}}{\partial \theta} \Big)
\end{aligned}
\end{equation}
Similarly, we take gradients to the $\mathcal L_{MS}$ loss function~(\ref{equ:Ms-ori}):
\begin{equation}
\label{equ:gradients-of-MS}
\begin{aligned}
\frac{\partial \mathcal L_{LS}}{\partial \theta} &= \sum_{i=1}^{B} \Big(\sum_{j \in \P_i}\frac{-1}{e^{\alpha(S_{ij}-\lambda)} + \sum\limits_{k\in\P_i}^B e^{\alpha(S_{ij} -S_{ik})}}  \frac{\partial S_{ij}}{\partial \theta} +\sum_{j \in \N_i} \frac{1}{e^{\beta(\lambda - S_{ij})} + \sum\limits_{k\in\N_i}^B e^{\beta(S_{ik} - S_{ij})}}    \frac{\partial S_{ij}}{\partial \theta} \Big)
\end{aligned}
\end{equation}
By substituting~(\ref{equ:closed-form-MS}) into~(\ref{equ:gradients-of-g-2}),and set $c^+ = \lambda + m, c^- = \lambda - m, \gamma_i^+ = \frac{1}{\alpha},  \gamma_i^- = \frac{1}{\beta}, i\in[B]$, it is obvious to show equation ~(\ref{equ:gradients-of-g-2}) and~(\ref{equ:gradients-of-MS}) are the same. As a result, our DRO framework also recovers the MS loss.

\subsubsection{Recovering of HAP2S\_E in~\cite{yu2018hard}}

\cite{yu2018hard} provides an a hardaware point to set (HAP2S) triplet loss with an adaptive hard mining scheme to address the sensitive issue caused by mining the hardest positive pair and negative pair in~\cite{hermans2017defense}. The key of HAP2S loss is to assign different
weights to the points in $\P_i$ and $\N_i$ for each anchor $\x_i$.
We show that our DRO framework is able to recover HAP2S with exponential weighting scheme by exactly using the DRO formulation equation~(\ref{equ:DRO_Lifted_Structure_Loss}) for LS loss.

The triplet loss is defined as follows:
\begin{equation}
    \begin{aligned}
    \mathcal{L}_{trp} = \frac{1}{N_t}\sum\limits_{y_j = y_i, y_k\neq y_i }^{}  [S_{ik}-S_{ij}+m]_+
    \end{aligned}
\end{equation}
where $[x]_+ = \max \{ 0, x \}$, $S_{ij}$ and $S_{ik}$ denote the similarity of positive pair $\{\x_i, x_j\}, j \in \P_{i}$, negative pair $\{\x_i, x_k\}, k\in \N_{i}$ for the same anchor $\x_i$, $N_t$ is the number of all possible triplets in the mini-batch.
Without causing ambiguity, $S_{ij}^+ = S_{ij}$ , when $j\in \P_i$, $S_{ij}^- = S_{ij}$, when $j \in \N_i$. 
Similar to pair losses, such as margin loss and binomial loss, there exist a huge amount of triplets in a batch that have no contribution to the $\mathcal L_{trp}$. As a result, pair mining is critical to improve the performance of the model. 

\cite{hermans2017defense} provide a variant of
triplet loss by selecting the hardest positive pair and the hardest negative pair for each anchor. The formulation can be written as: 
\begin{equation}
\label{equ:BH}
    \begin{aligned}
    \mathcal{L}_{trpBH} = \frac{1}{B}\sum\limits_{ i=1}^{B}  [\max\limits_{j\in \N_i}S_{ij}-\min\limits_{j\in \P_i}S_{ij}+m]_+
    \end{aligned}
\end{equation}
The state-of-the-art results on two large-scale datasets has been reported based on this variant of triplet loss.
However, equation~(\ref{equ:BH}) is sensitive to outliers which usually serve as the hardest sample. 
To increase the robustness of the model, HAP2S triplet loss has been proposed in~\cite{yu2018hard}:
\begin{equation}
\label{equ:HAP2S}
    \begin{aligned}
    \mathcal{L}_{HAP2S} = \frac{1}{B}\sum\limits_{ a=1}^{B}  [\mathcal{S}_{ij}^- -\mathcal{S}_{ij}^+ +m]_+   ,
    \end{aligned}
\end{equation}
where
\begin{equation}
\label{equ:HAP2S_S}
\begin{aligned}
\mathcal{S}_{ij}^- = \frac{\sum\limits_{k\in \N_i}q_{ij}^- S_{ij}^-}{\sum\limits_{j\in \N_i} q_{ij}^-}, \mathcal{S}_{ij}^+ = \frac{\sum\limits_{j\in \P_i}q_{ij}^+ S_{ij}^+}{\sum\limits_{j\in \P_i} q_{ij}^+},
\end{aligned}
\end{equation}
where $q_{ij}$ is the weights designed to each pair,
for which they propose two variants of weighting schemes for HAP2S, i.e., exponential weighting and polynomial weighting. 
Here we show that our DRO formulation is able to recover the HAP2S with exponential weighting scheme (denoted by HAP2S\_E), i.e. the weight $q_{ij}$ of each pair is an exponential function over its similarity:
\begin{equation}
\label{equ:q}
\begin{aligned}
q_{ij}^+ = \exp{ ( \frac{ -S_{ij} }{ \gamma } ) }, j \in \P_i , ~~
q_{ij}^- = \exp{ ( \frac{  S_{ij} }{ \gamma } ) }, j \in \N_i .
\end{aligned}
\end{equation}
Note that $q_{ij}^+$ and $q_{ij}^-$ are constant scalars that not involved in the gradient of $\mathcal{L}_{HAP2S}$ w.r.t to $\theta$.

Substitute equation~(\ref{equ:HAP2S_S}) and (\ref{equ:q}) into~(\ref{equ:HAP2S})
\begin{equation}
    \begin{aligned}
    \mathcal{L}_{HAP2S} = \frac{1}{B}\sum\limits_{
    i=1}^{B}  \Big( \sum\limits_{j\in \P_i}(\frac{-1}{\sum\limits_{k\in\P_i}\exp(\frac{S_{ij}-S_{ik}}{\gamma})}) {S}_{ij}+\sum\limits_{j\in \N_i}(\frac{1}{\sum\limits_{k\in\N_i}\exp(\frac{S_{ik}-S_{ij}}{\gamma})}) {S}_{ij}\Big) + D 
    \end{aligned}
\end{equation}
where $D$ absorbs all constants.

Since $S_{ij}-S_{ik}$, $S_{ik}-S_{ij}$ in the exponential function is derived from $q_{ij}^+$ and $q_{ij}^-$, thus they are also a constant scalars when taking derivative to $\theta$.
By taking derivative to $\mathcal L_{HAP2S}$, we get:
\begin{equation}
\label{equ:gradient-form-HAP2S}
\begin{aligned}
\frac{\partial \mathcal{L}_{HAP2S}}{\partial\theta} = \frac{1}{B}\sum\limits_{
    i=1}^{B}  \Big( \sum\limits_{j\in \P_i}(\frac{-1}{\sum\limits_{k\in\P_i}\exp(\frac{S_{ij}-S_{ik}}{\gamma})})
    \frac{\partial {S}_{ij}}{\partial \theta} +\sum\limits_{j\in \N_i}(\frac{1}{\sum\limits_{k\in\N_i}\exp(\frac{S_{ik}-S_{ij}}{\gamma})}) \frac{\partial{S}_{ij}}{\partial \theta} \Big),
\end{aligned}
\end{equation}
We can see that the subgradients of $\mathcal{L}_{HAP2E} $  in ~(\ref{equ:gradient-form-HAP2S}) are the same as the subgradients of LS loss in~(\ref{equ:gradients-of-LS}).  
As a result, $\mathcal{L}_{HAP2S}$ is a special case of our DRO framework.

\subsection{Supplementary Experiments}

\begin{figure}[h]
    \centering
    \begin{minipage}[t]{0.49\textwidth}
    \includegraphics[width=\textwidth]{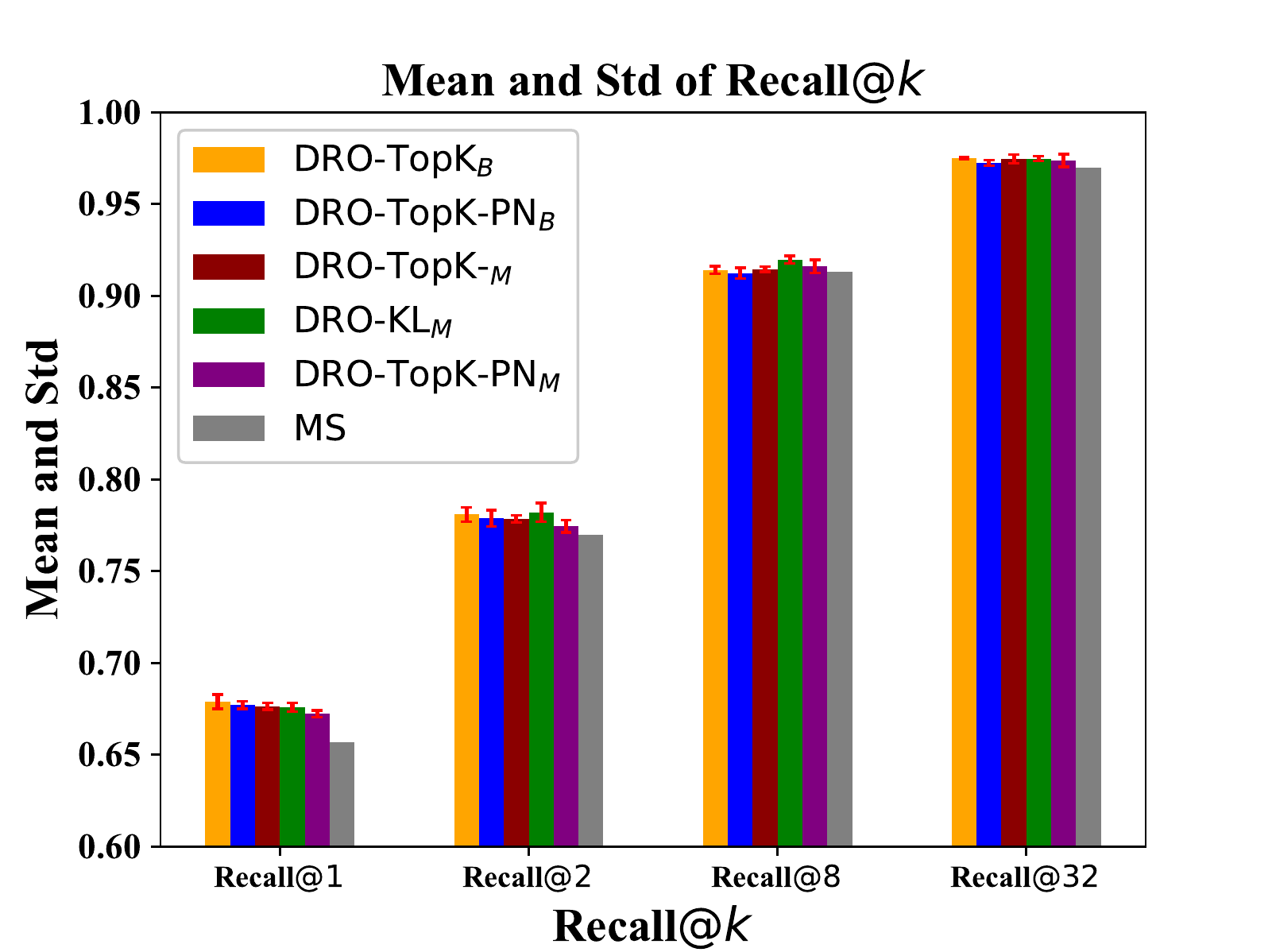}
   \vspace{-0.8\baselineskip}
    \caption{Mean and std of Recall@k over five runs comparing the best baseline performer (MS)}
    \label{fig:mean_and_std}
    \end{minipage}
    \begin{minipage}[t]{0.49\textwidth}
    \includegraphics[width =\textwidth]{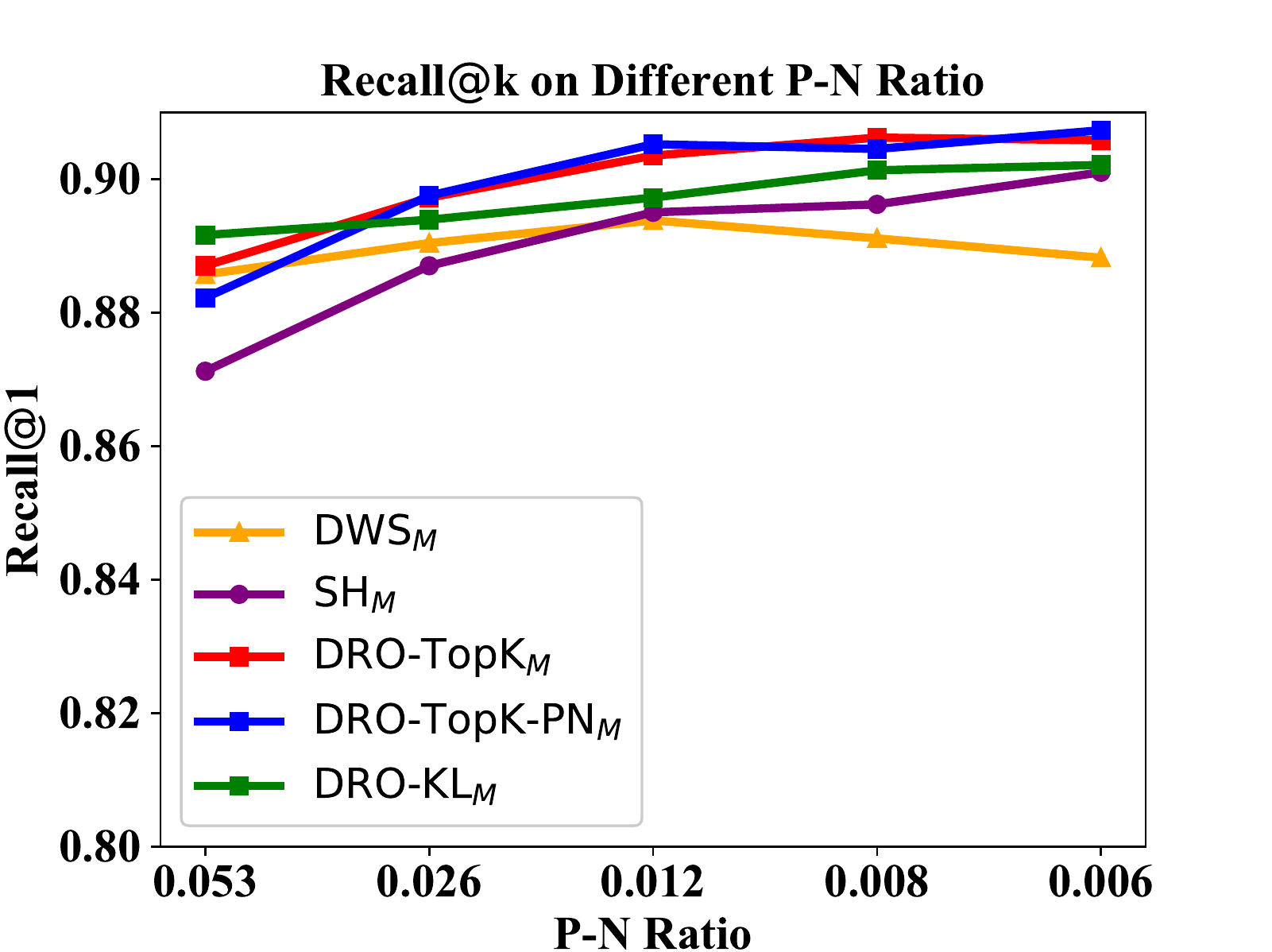}
    \vspace{-0.8\baselineskip}
    \caption{Recall@1 vs Imbalance Ratio on embedding space dimension 512}
    \label{fig:runtime-imbanlance-512}
    \end{minipage}
\end{figure}

To investigate the effect of randomness of the stochastic algorithms and evaluate the robustness of our DRO framework, we report the average mean and standard variance of recall$@k$ on Cub-200-2011 in Figure~\ref{fig:mean_and_std}. 
We do not plot recall$@4$ and recall$@16$ for better visualization. 
The experimental setting is the same as the experiments of the SOTA quantitative results we reported in Table~\ref{tab:SOTA_cub_car} (section~\ref{sec:Quantitative Results}) but with five runs. 
The gray bars are the recall$@k$ of best performer among SOTA baselines, i.e., MS.
It is clear to see that all our DRO variants outperform MS in terms of the average recall$@k$ over all different values of $k$. Specifically, the average recall$@1$ of DRO-TopK$_B$ is $67.9\%$, which has a significant improvement over the baselines, i.e., $65.7\%$ of MS. 
In addition, the small standard deviation error bars imply that our DRO framework is robust enough to have a better performance than SOTA methods.

To show the effect of the network architecture to our DRO framework and its robustness, we additionally repeat the experiments of Recall/Imbalance Ratio in section~\ref{sec:Capacity to Handle Pair Imbalance.}, but with the embedding space dimensions $512$ (rather than 1024 in Section~\ref{sec:Capacity to Handle Pair Imbalance.}). The results are illustrated in Figure~\ref{fig:runtime-imbanlance-512}. In comparison with the performance of recall$@1$ on embedding space dimensions 1024 in Figure~\ref{fig:resisting_imbalance}, we can see that the fluctuations of recall$@1$ on different PN-Ratios are sublet when the feature embedding changes from 1024 to 512.
For example, the recall$@1$ only changes, from 0.9046 to 0.9058 for DRO-TopK$_M$, from 0.9086 to 0.9073 for DRO-TopK-PN$_{M}$, from 0.9018 to 0.9021 for DRO-KL$_M$, when PN-Ratio is 0.006. 
Further, the trends of different methods imply that our DRO framework can consistently achieve competitive even best results compared with other SOTA methods (SH and DWS) under different embedding space dimensions. 
Since different P-N Ratios in corresponding to batchsizes $\{80, 160, 320, 480, 640 \}$, the above experimental results verify that our methods are not too sensitive to the embedding space dimensions in different batchsizes, and also outperform other SOTA mining methods in different embedding space dimensions. 
% \textcolor{red}{
In contrast, it shows that the baseline DWS is relatively more sensitive to the embedding space dimension. 
In Figure \ref{fig:runtime-imbanlance-512} where embedding space dimension is set to $512$, DWS has a smaller performance drop from P-N ration $=0.012$ to $0.006$,
while in Figure \ref{fig:resisting_imbalance} where the embedding space dimension is $1024$, DWS encounters a much larger performance decrease.
% }

%Then, from the perspective of (\ref{equ:Closed-form-Weighting}), the above equation implies:
%\begin{equation}
%\begin{aligned}
%w_{ij}^- & =\frac{1}{\sum\limits_{k\in\N_i}\exp(\frac{S_{ik}-S_{ij}}{\gamma})} \\   
%w_{ij}^+ &= \frac{1}{\sum\limits_{k\in\P_i}\exp(\frac{S_{ij}-S_{ik}}{\gamma})}
%\end{aligned}
%\end{equation}
%which, when $\P_i = \P_i, \N_i = \N_i$, are exactly the same as $p_{ij}^{+*}$, $p_{ij}^{-*}$ in equation~(\ref{equ:close-form-LS}).
%Which means that under our DRO framework, HAP2S\_E is equivalent to LS loss, as in (\ref{equ:DRO_Lifted_Structure_Loss}).
\end{document}

%% file: math_commands.tex
\usepackage{amsmath,amsfonts,bm}

\def\eqref#1{equation~\ref{#1}}
\def\1{\bm{1}}

\DeclareMathAlphabet{\mathsfit}{\encodingdefault}{\sfdefault}{m}{sl}
\SetMathAlphabet{\mathsfit}{bold}{\encodingdefault}{\sfdefault}{bx}{n}